\documentclass[a4paper,10pt]{article}

\usepackage[UKenglish]{babel}
\usepackage{graphicx}
\usepackage{subcaption}
\usepackage{amsmath}
\usepackage{amssymb}
\usepackage{booktabs}
\usepackage{multirow}
\usepackage{enumitem}
\usepackage{geometry}
\usepackage{fancyhdr}
\usepackage{color}
\usepackage[table]{xcolor}
\usepackage{makecell}
\usepackage{cleveref}
\usepackage{lineno}
\usepackage{setspace}
\usepackage{authblk}
\usepackage{parskip}
\usepackage[style=nature,sorting=none,autocite=superscript]{biblatex}
\usepackage{titlesec}
\usepackage{csquotes}
\usepackage{verbatim}
\usepackage{totcount}
\usepackage{pdfpages}

\crefname{figure}{Fig.}{Figs.}
\Crefname{figure}{Fig.}{Figs.}
\crefname{table}{Table}{Tables}
\Crefname{table}{Table}{Tables}

\titleformat*{\section}{\large\bfseries}
\titleformat*{\subsection}{\normalsize\bfseries}
\titleformat*{\subsubsection}{\normalsize\itshape}

\DeclareCaptionLabelSeparator{bar}{ $|$ }
\regtotcounter{figure}
\regtotcounter{table}
\newtotcounter{citnum}
\AtEveryBibitem{\stepcounter{citnum}}

\setlist{noitemsep}

\begin{document}

\pagenumbering{gobble}
\begin{titlepage}

\newgeometry{bottom=1cm,right=2cm}

\textbf{\large Enabling clinical use of foundation models for computational pathology}

\vspace{0.5cm}

Audun L. Henriksen\textsuperscript{1*},
Ole-Johan Skrede\textsuperscript{1*},
Lisa van der Schee\textsuperscript{2},
Enric Domingo\textsuperscript{3,4},
Karolina Cyll\textsuperscript{1},
Sepp De Raedt\textsuperscript{1},
Ilyá Kostolomov\textsuperscript{1},
Jennifer Hay\textsuperscript{5},
Wanja Kildal\textsuperscript{1},
Joakim Kalsnes\textsuperscript{1},
Robert W. Williams\textsuperscript{6},
Manohar Pradhan\textsuperscript{1},
John Arne Nesheim\textsuperscript{1},
Hanne A. Askautrud\textsuperscript{1},
Maria X. Isaksen\textsuperscript{1},
Karmele Saez de Gordoa\textsuperscript{7,8,9},
Miriam Cuatrecasas\textsuperscript{7,8,9},
Joanne Edwards\textsuperscript{10},
TransSCOT group\textsuperscript{‡},
Arild Nesbakken\textsuperscript{11,12},
Neil A. Shepherd\textsuperscript{13},
Ian Tomlinson\textsuperscript{3},
Daniel-Christoph Wagner\textsuperscript{14},
Rachel S. Kerr\textsuperscript{3},
Tarjei Sveinsgjerd Hveem\textsuperscript{1},
Knut Liestøl\textsuperscript{1},
Yoshiaki Nakamura\textsuperscript{3,15,16},
Marco Novelli\textsuperscript{17},
Masaaki Miyo\textsuperscript{18},
Sebastian Foersch\textsuperscript{14},
David N. Church\textsuperscript{19,20},
Miangela M. Lacle\textsuperscript{2},
David J. Kerr\textsuperscript{21},
Andreas Kleppe\textsuperscript{†1,22,23}\\

\vspace{-0.1cm}

\textsuperscript{1}Institute for Cancer Genetics and Informatics, Oslo University Hospital, Oslo, Norway\\
\textsuperscript{2}Department of Pathology, University Medical Center Utrecht, Utrecht, The Netherlands\\
\textsuperscript{3}Department of Oncology, University of Oxford, Oxford, UK\\
\textsuperscript{4}CRUK Beatson Institute of Cancer Research, Garscube Estate, Glasgow, UK\\
\textsuperscript{5}Glasgow Tissue Research Facility, University of Glasgow, Queen Elizabeth University Hospital, Glasgow, UK\\
\textsuperscript{6}Area for Improvement and Digital Transformation, Norwegian Offshore Directorate, Stavanger, Norway\\
\textsuperscript{7}Pathology Department, Hospital Clínic, Barcelona, Spain\\
\textsuperscript{8}Institut d’Investigacions Biomèdiques August Pi I Sunyer (IDIBAPS), Barcelona, Spain\\
\textsuperscript{9}Department of Clinical Foundations, Universitat de Barcelona, Barcelona, Spain\\
\textsuperscript{10}School of Cancer Sciences, Wolfson Wohl Cancer Research Centre, University of Glasgow, Glasgow, UK\\
\textsuperscript{11}Institute of Clinical Medicine, University of Oslo, Oslo, Norway\\
\textsuperscript{12}Department of Gastrointestinal Surgery, Oslo University Hospital, Oslo, Norway\\
\textsuperscript{13}Gloucestershire Cellular Pathology Laboratory, Cheltenham General Hospital, Cheltenham, UK\\
\textsuperscript{14}Institute of Pathology, University Medical Center, Mainz, Germany\\
\textsuperscript{15}Department of Gastrointestinal Oncology, National Cancer Center Hospital East, Kashiwa, Japan\\
\textsuperscript{16}Translational Research Support Office, National Cancer Center Hospital East, Kashiwa, Japan\\
\textsuperscript{17}Department of Cancer Biology and Pathology, University College London, London, UK\\
\textsuperscript{18}Department of Gastroenterological Surgery, Osaka International Cancer Institute, Osaka, Japan\\
\textsuperscript{19}Centre for Human Genetics, University of Oxford, Oxford, UK\\
\textsuperscript{20}Oxford NIHR Comprehensive Biomedical Research Centre, Oxford University Hospitals NHS Foundation Trust, John Radcliffe Hospital, Oxford, UK\\
\textsuperscript{21}Nuffield Division of Clinical Laboratory Sciences, University of Oxford, Oxford, UK\\
\textsuperscript{22}Department of Informatics, University of Oslo, Oslo, Norway\\
\textsuperscript{23}Centre for Research-based Innovation Visual Intelligence, UiT The Arctic University of Norway, Tromsø, Norway\\

\vspace{-0.1cm}

\textsuperscript{*}Joint first authors with equal contribution.\\

\vspace{-0.2cm}

\textsuperscript{†}Corresponding author:\\
Andreas Kleppe\\
Institute for Cancer Genetics and Informatics, Oslo University Hospital\\
NO-0424 Oslo, Norway\\
andrekle@ifi.uio.no\\

\vspace{-0.2cm}

\textsuperscript{‡}The individual names of the TransSCOT Trial Management Group are listed in the Appendix to the main manuscript, and the individuals should be listed as collaborators.\\

\restoregeometry%
\end{titlepage}

\doublespacing%

\clearpage
\section{Abstract}

Foundation models for computational pathology are expected to facilitate the development of high-performing, 
generalisable deep learning systems. However, in addition to biologically relevant features, current foundation models also capture pre-analytic and scanner-specific variation that bias the predictions made by downstream task-specific models trained on these features.
Here we show that introducing novel robustness losses during downstream model training reduces sensitivity to technical variability. A purpose-designed comprehensive experimentation setup with 27,042 whole-slide images from 6,155 patients is used to train thousands of models from the features of eight well-known foundation models for computational pathology. In addition to a substantial improvement in robustness, our approach improves classification accuracy by focusing on biologically relevant features. 
It mitigates robustness limitations of foundation models for
computational pathology without retraining the foundation models themselves, enabling development of models that are more suitable in real-world clinical use.

\clearpage
\pagenumbering{arabic}  %
\section{Introduction}

Inspired by the success of foundation models in natural language processing, several
pathology foundation models have been published in recent
years.\autocite{bommasani2021opportunities,chen2024towards,xu2024whole, Vorontsov2024AFM,zimmermann2024virchow2,filiot2024phikon,nechaev2024hibou}
These models are trained in a self-supervised manner on large and diverse datasets of whole-slide images (WSIs) to learn general-purpose feature representations that can be reused for a variety of tasks. Smaller downstream task-specific models can be built on these pretrained features for tissue classification, biomarker assessment and survival prediction.\autocite{da2026computationalpathologyeraemerging} This is particularly attractive in computational pathology, where annotated datasets are often limited and developing deep learning models can be time-consuming and costly.

Deep learning models do not inherently distinguish causal biological features from spurious correlations present in the training data and may exploit them to optimise performance, a phenomenon known as shortcut learning.\autocite{geirhos2020, Lapuschkin2019UnmaskingCH}
In medical imaging, this is a prominent issue, as numerous sources of variation in acquisition procedures and equipment can introduce biologically irrelevant patterns that models may exploit. Well-known examples include surgical markings in dermatology,\autocite{narla2018,winkler2019association} radiographic markers in X-ray images,\autocite{zech2018}
and institutional signatures embedded in histological images.\autocite{howard2021,dehkharghanian2023,Clusmann2025IncidentalPI} 
In histopathology, such patterns can result from variation in tissue processing, \autocite{howard2021,kheiri2025} staining protocols, \autocite{tellez2019,lin2025} and digital scanning equipment.\autocite{Stacke2020MeasuringDS,aubreville2023mitosis,thiringer2026}
When models rely on such technical rather than biological features, their performance may drop on data acquired under different conditions, limiting reliability in clinical practice.\autocite{Kleppe2021DesigningDL,vanderLaak2021DeepLI}

Interest in computational pathology foundation models is partly driven by the expectation that large-scale pretraining will improve the robustness and generalisability of downstream models. However, recent studies indicate that foundation models themselves remain sensitive to technical variation. Evaluations of  
multiple pathology foundation models show that their features capture medical centre and scanner attributes more strongly than underlying biological characteristics, regardless of training data scale or model size.\autocite{dejong2025, thiringer2026, komen2024do, komen2025towards} Moreover, downstream models built on foundation models do not consistently achieve strong performance,\autocite{gustafsson2024evaluating, mulliqi2025, Neidlinger2025BenchmarkingFM} and, when sufficient task-specific data are available, they do not necessarily outperform conventional approaches.\autocite{mulliqi2025} Efforts to improve downstream robustness have largely targeted technical variation either at the image level or in features extracted from small image patches (tiles), with only modest and inconsistent gains in robustness and downstream performance.\autocite{komen2025towards,mishra2025comparing, komen2024do, Huang2025KnowledgeguidedAO,komen2025towards,carloni2025,Ryu2025SCORPIONAS}

We propose a training framework that mitigates the robustness problem at the level of the downstream model by encouraging consistent predictions for the same tissue region in scanner-paired WSIs. This is achieved by regularising the task-specific network with robustness loss terms applied to co-registered tiles. 
By enforcing these constraints throughout the model, robustness improves not only in tile-level representations but also in the final WSI predictions. We demonstrate utility of this regularisation through extensive experimentation with eight well-known foundation models for computational
pathology and two clinical tasks in colorectal cancer (CRC); survival outcome prediction
and lymph node metastasis (LNM) prediction in pathological T1 (pT1) CRC. Multiple WSIs of the same hematoxylin and eosin (H\&E)-stained tissue section scanned on different scanners were included, as well as different tissue sections of the same tumour that were prepared and imaged in different laboratories. In total, 27,042 WSIs from 6,155 patients were used to train and externally test thousands of downstream task-specific models based on systematic tuning of about 350,000 models.
We show that the proposed framework substantially improves robustness to scanner
variation across all models and additionally improves prediction accuracy through focusing on biologically relevant features.

\clearpage
\section{Results}

\subsection{Foundation models for computational pathology are sensitive to non-biological differences}

\begin{figure}[ht]
  \centering
  \includegraphics[width=\textwidth]{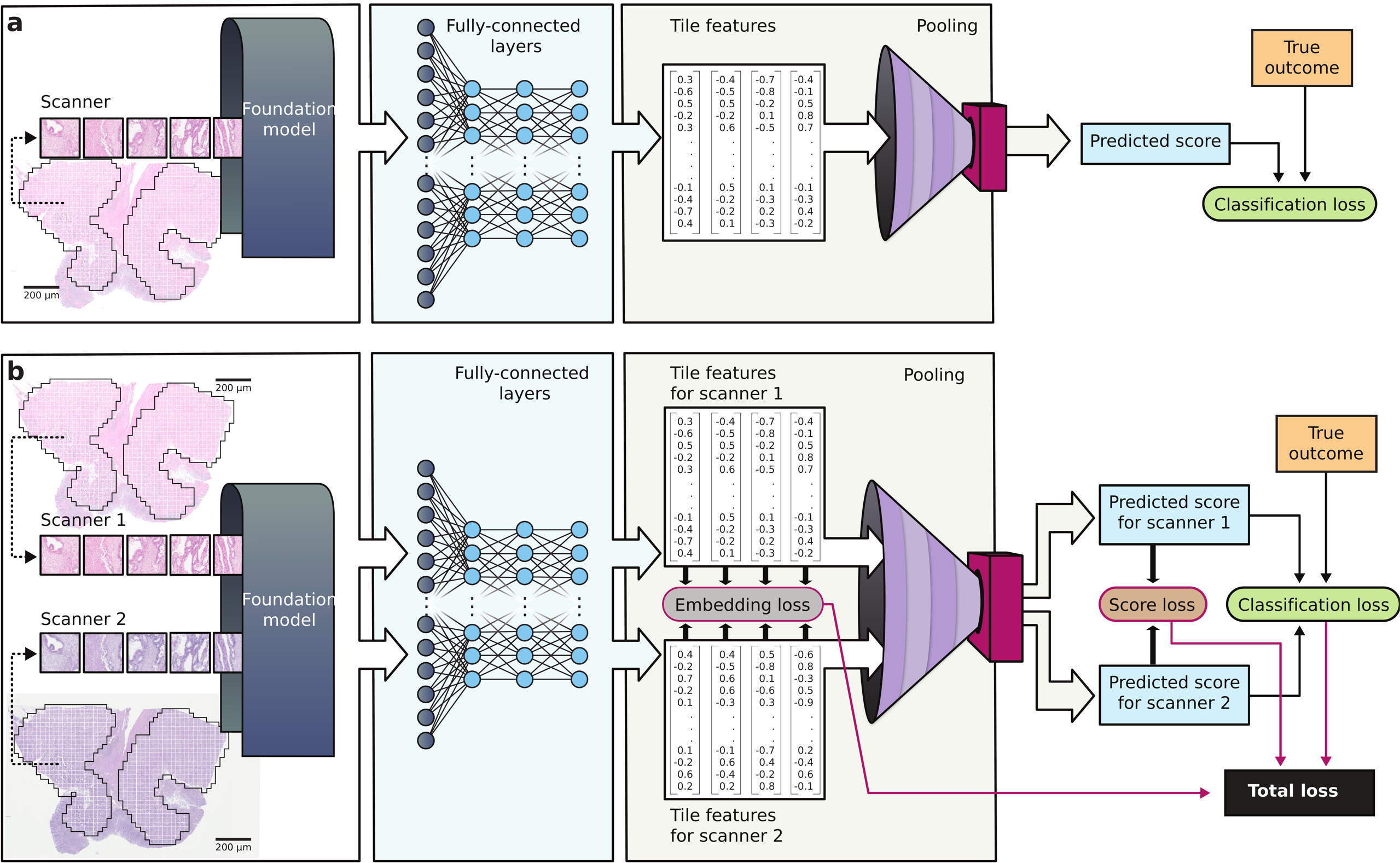}
  \caption[Method overview]{%
    \textbf{Method overview}\\
  }\label{fig:method}
\end{figure}

\begin{figure}[ht]
  \centering
  \includegraphics[width=\textwidth]{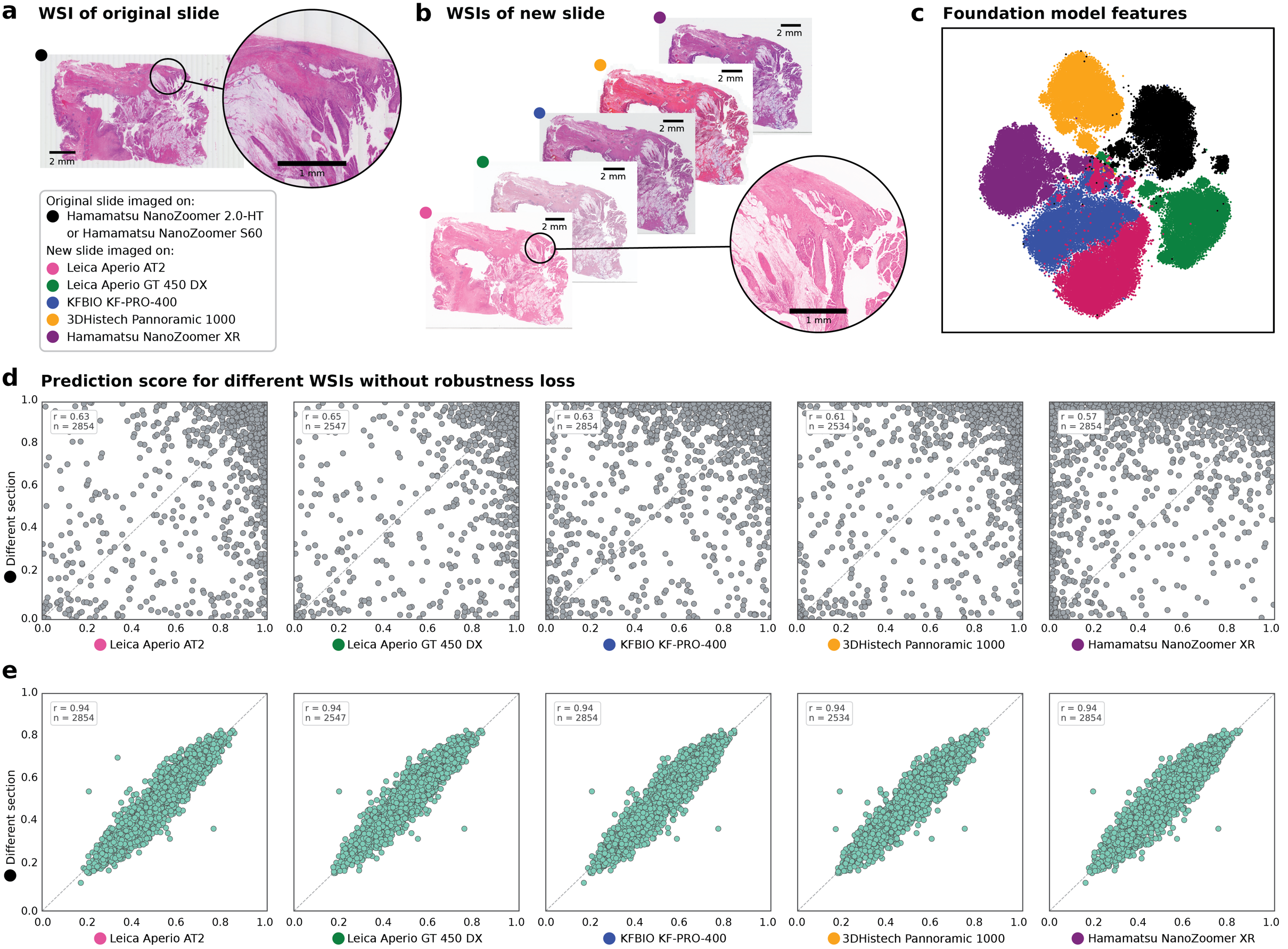}
  \caption[Foundation model robustness issue]{%
    \textbf{Foundation model robustness issue}\\
  }\label{fig:scan-comparison}
\end{figure}

Features from 7,777 WSIs from 2,250 patients are extracted using published foundation models: H-Optimus-0\autocite{hoptimus0}, H-Optimus-1\autocite{hoptimus1}, Hibou-L\autocite{nechaev2024hibou}, Phikon-v2\autocite{filiot2024phikon}, Prov-GigaPath\autocite{xu2024whole}, UNI, UNI2\autocite{chen2024towards} and Virchow2\autocite{zimmermann2024virchow2, Vorontsov2024AFM}.
First, we use these features to train and tune deep learning models for CRC survival outcome prediction using a common attention-based multiple instance learning approach (\cref{fig:method}a)\autocite{ilse2018attention}.
To test the robustness to sample preparation and imaging variation of the resulting models, we analysed two tissue slides from each of 2,875 patients from TransSCOT, the translational arm of the SCOT trial\autocite{Iveson20183V6}. For each patient, one slide consisted of a 4~\textmu m thick section that was sectioned, stained and imaged in UK (hereafter named `original slide'), while the other consisted of a 5~\textmu m thick section that was stained and imaged on five different pathology scanners in Norway (\cref{fig:scan-comparison}a,b).

Mapping the high-dimensional features from a foundation model into two-dimensional points with t-distributed stochastic neighbour embedding (t-SNE)\autocite{van2008visualizing} and labelling them with the source of origin clearly shows that the foundation model features cluster WSIs primarily by origin and not biological differences (\cref{fig:scan-comparison}c).
Comparison of prediction scores produced by the conventionally trained models for different WSIs from the same patient showed
substantial variability related to the laboratory site and scanner device for all foundation models (\cref{fig:scan-comparison}d and Extended Data \cref{e-fig:scatterplots_KDE}).
These findings indicate that conventionally trained downstream models inadvertently apply non-biological information contained in the features from foundation models, resulting in patient representations that are suboptimal for the intended prediction task. 

By quantifying robustness as the relative average standard deviation of the prediction score for different WSIs from the same patient, referred to as the inconsistency metric (see \emph{Methods} for more details), we find that the inconsistency varies between foundation models as well as between
individual training runs using features from the same foundation model (\cref{fig:outcome-comparison-boxplot}b, Extended Data \cref{e-tab:test-outcome-improvement}).
Hibou-L and Phikon-v2 show the highest average inconsistency of 0.65 and 0.54,
respectively.
Prov-GigaPath, Virchow2, H-Optimus-0, H-Optimus-1, UNI and UNI2 display lower
inconsistency, ranging from 0.31 to 0.47.

After dichotomising the prediction scores into classifications of patient outcome by selecting the class with highest score, we measure the percentage of patients that are not classified differently in a pair of different WSI variants (different slide or scanner), averaged over all pairs with different WSI variants. Models trained on Hibou-L features have an
average classification agreement of about 75\%, Phikon-v2 models have about 82\% and models trained on features from one of the other six
foundation models have between 85\% and 91\% (\cref{fig:outcome-comparison-boxplot}c, Extended Data \cref{e-tab:test-outcome-improvement}).

\subsection{Generalisation of scanner information in foundation model features}

To investigate the consistency across datasets for the scanner information contained within the foundation model features, we trained a simple linear classifier to classify the originating scanner directly from the foundation model features using the QUASAR 2 dataset and externally test the classifier on the TransSCOT dataset. This linear probing shows that all five scanners can be accurately identified across datasets, with an average classification accuracy in the external test dataset of 1.000 for Phikon-v2, above 0.998 for H-Optimus-0, H-Optimus-1, UNI, UNI2 and Virchow2, 0.992 for Prov-GigaPath and 0.986 for Hibou-L (Extended Data \cref{e-fig:linear-probing}). This indicates that non-biological information is consistently encoded across patients and sample preparations, highlighting the prominence of these attributes in foundation model features.

\subsection{Training robust downstream task-specific models from foundation model features}

To encourage similar predictions for different WSIs of the same tissue section, we include two additional loss terms; a contrastive loss on tile-level embeddings, based on InfoNCE\autocite{oord2018representation}, that pulls together features of the same physical area imaged on different scanners while pushing apart features from different patients, and a mean squared error loss on the prediction scores that penalises differences in the final slide-level predictions between corresponding co-registered WSIs (\cref{fig:method}b). Each of these loss terms are multiplied with a weight factor and added to the ordinary classification loss to define the full loss function.

\begin{figure}[ht]
  \centering
  \includegraphics[width=\textwidth]{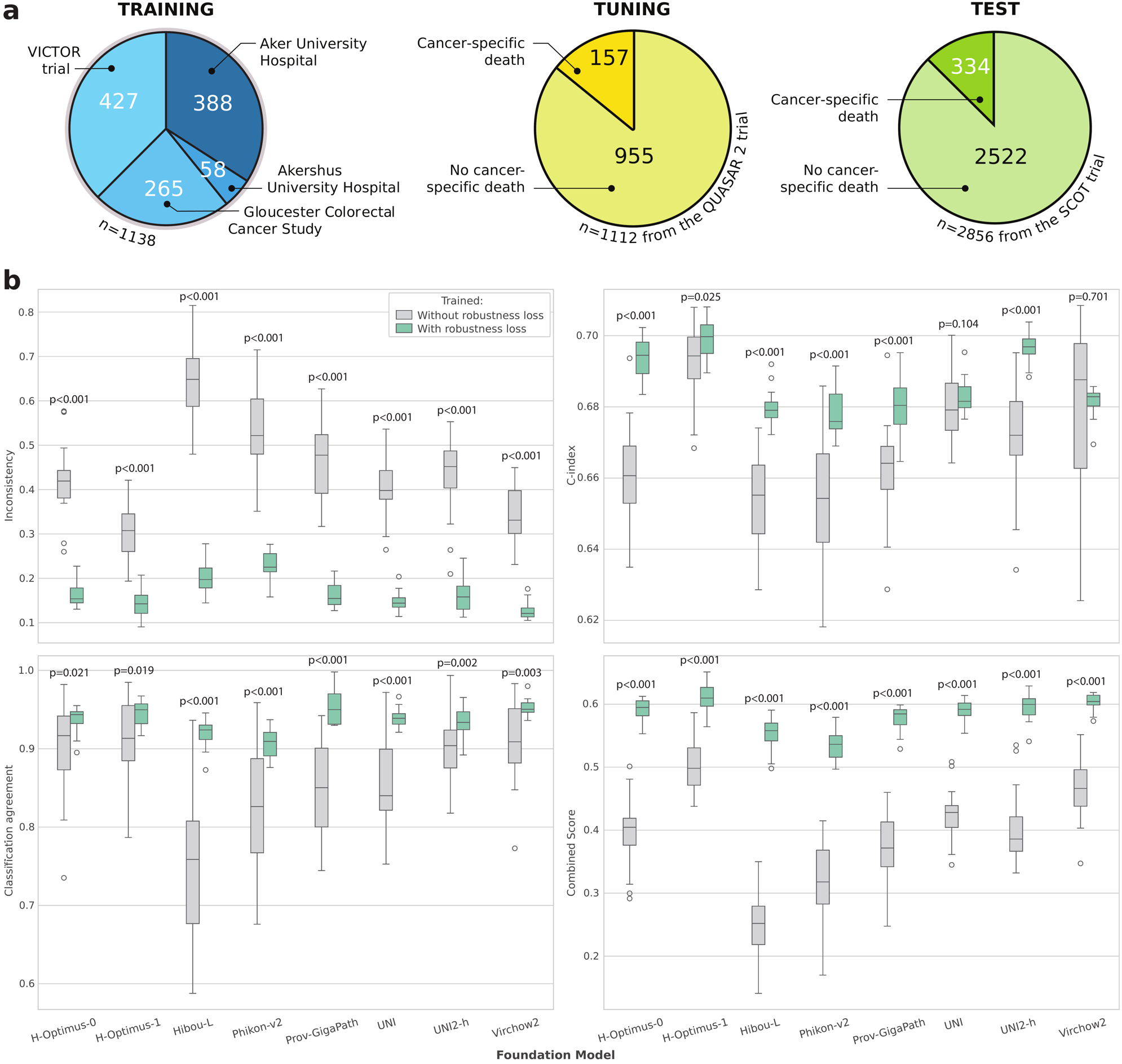}
  \caption[Survival prediction]{%
    \textbf{Survival prediction}\\
  }\label{fig:outcome-comparison-boxplot}
\end{figure}

Using these additional loss terms, we train and tune downstream task-specific models to predict survival outcomes from foundation model features extracted from 7,777 WSIs from 2,250 patients (\cref{fig:outcome-comparison-boxplot}a).
In the TransSCOT test dataset, the prediction scores from different WSIs of the same patient are markedly more similar when robustness losses are applied than in models trained conventionally without these losses (\cref{fig:scan-comparison}d,e and Extended Data \cref{e-fig:scatterplots_KDE}).
Consistent with this, multiple robustness metrics improve (\cref{fig:outcome-comparison-boxplot} and Extended Data \cref{e-tab:test-outcome-improvement}).
Average inconsistency decreases to below 0.17 for all foundation models except Hibou-L and Phikon-v2, which show average inconsistency values of 0.20 and 0.23, respectively  
(\cref{fig:outcome-comparison-boxplot}b).
Relative to conventional training, these reductions are significant for all foundation models (all p < 0.001), with improvements ranging from 116\% to 217\%.
Average classification agreement also increases substantially (all p < 0.001), exceeding 93\% for all foundation models except Phikon-v2 and Hibou-L, with relative improvement from 50\% to 215\%. For all foundation models except Virchow2, the c-index also improves significantly with the addition of the robustness loss terms (\cref{fig:outcome-comparison-boxplot}b). Concordance correlation coefficient (CCC), a measure of agreement between paired scores,\autocite{lin1989}  show a similar pattern, with only these two models remaining below 96\% when robustness losses are applied and all models improving by 732\% to 1,397\% relative to conventional training (Extended Data \cref{e-tab:test-outcome-improvement}).

Models trained without the robustness losses frequently produce uniformly high prediction scores that are compressed into a narrow range and do not reflect true certainty.
Incorporating the robustness losses results in a broader and more intuitive spread of predictions, suggesting improved calibration and more clinically interpretable confidence estimates (\cref{fig:scan-comparison}d,e and Extended Data \cref{e-fig:scatterplots_KDE}).

\subsection{Congruity of spatially resolved predictions}

\begin{figure}[ht]
  \centering
  \includegraphics[width=\textwidth]{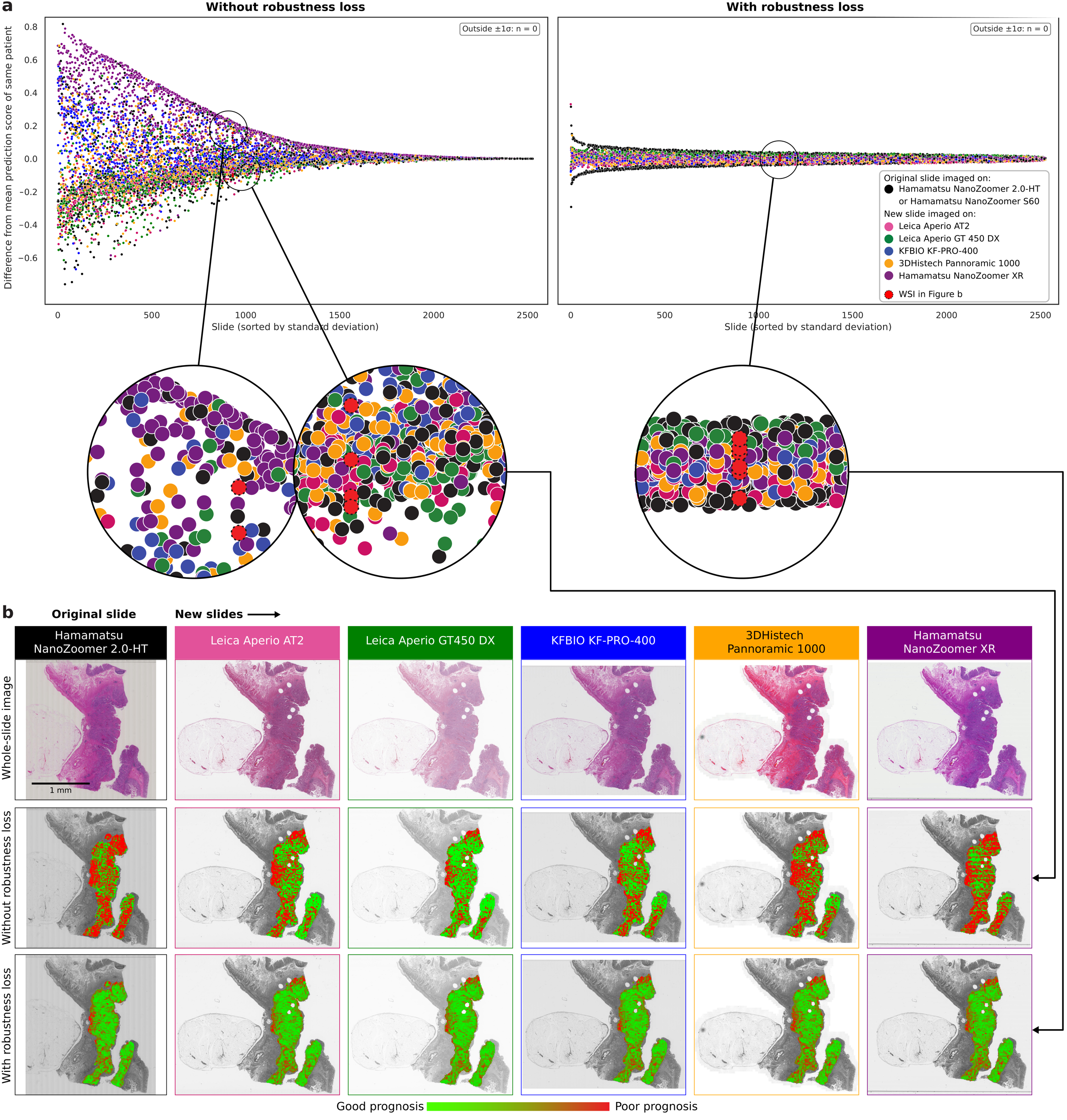}
  \caption[Spatial robustness]{%
    \textbf{Spatial robustness}\\
  }\label{fig:heatmap}
\end{figure}

The increased consistency of prediction scores for different WSIs from the same patient (\cref{fig:heatmap}a) indicates that the downstream task-specific model trained with robustness loss is less sensitive to non-biological variation in foundation model features related to sample preparation and scanner device. If this is true, then applying downstream task-specific models to individual image tiles should also produce more consistent predictions when the model is trained with robustness loss than when it is trained without it. Using a representative patient from the test dataset,  heatmaps of tile-level prediction scores demonstrate greater consistency between WSIs from the same patient when training the model with robustness loss (\cref{fig:heatmap}b).

\subsection{Analyses of each layer in the downstream task-specific models}

\begin{figure}[ht]
  \centering
  \includegraphics[width=\textwidth]{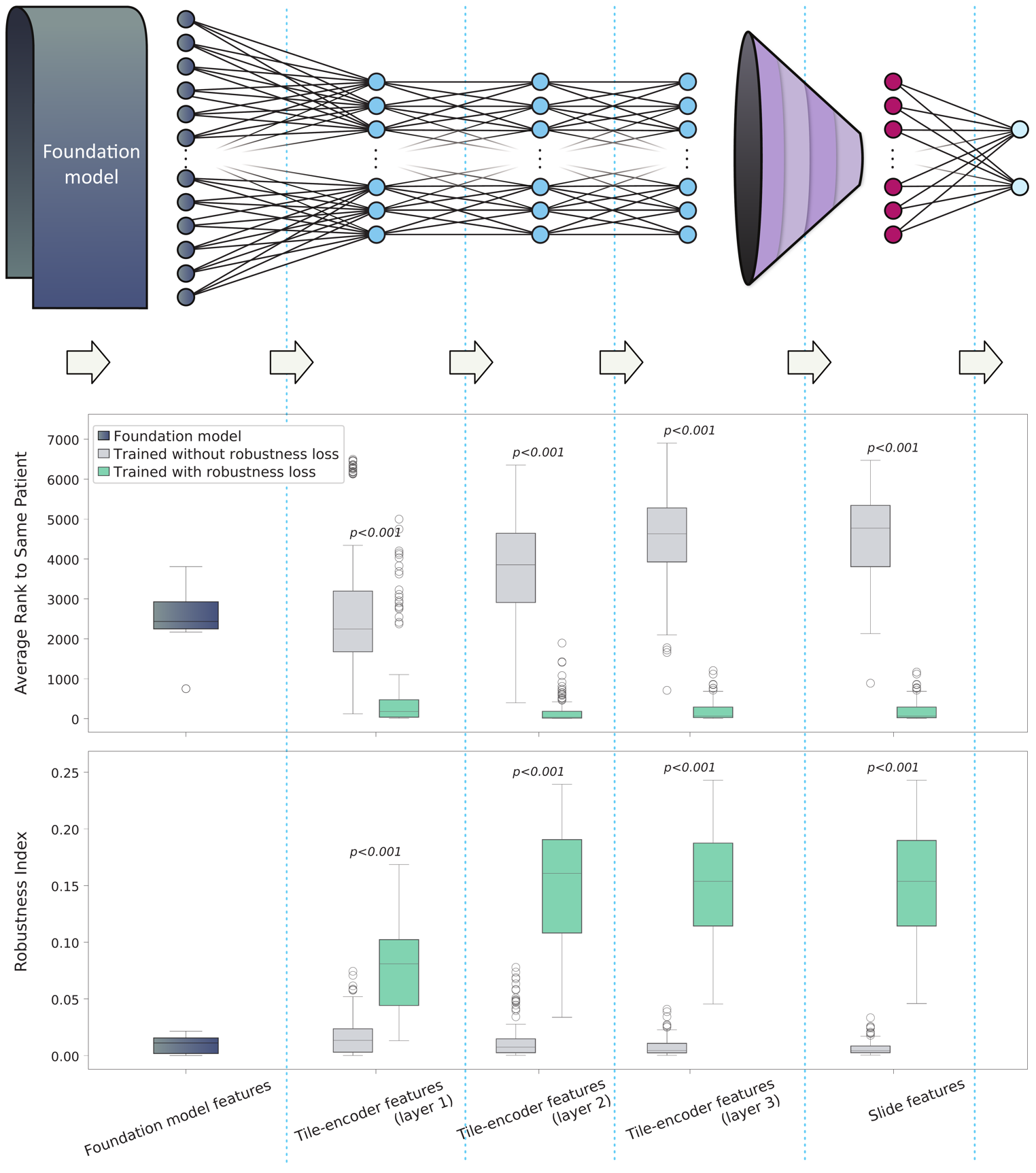}
  \caption[Robustness in each layer]{%
    \textbf{Robustness in each layer}\\
  }\label{fig:cosine}
\end{figure}

To understand where in the downstream task-specific models the robustness terms introduce changes, we analyse the features at each layer in the models. First, we compute Average Rank to Same Patient (ARSP) (see \emph{Methods} for more details) between WSIs from the same patient and find that it is statistically significantly smaller in each layer of the downstream task-specific models (\cref{fig:cosine} and Supplementary Fig. 10).
This shows that already the first layer of the downstream task-specific models learns to focus on biological information contained in the foundation model features, and ARSP also continues to decrease with subsequent layers of the model (\cref{fig:cosine} and Supplementary Fig. 10).
When analysing the Robustness Index (see \emph{Methods} for more details), we find that the features from models trained with robustness loss have statistically significant higher Robustness Index (\cref{fig:cosine} and Supplementary Fig. 10). This indicates that biological information is more dominant in the features from the models trained with robustness loss than those from the models trained without robustness loss, which suggests that the unique biology of each patient is more distinctly and better characterised when applying the robustness losses during training.

\subsection{Replication in a different prediction task}

\begin{figure}[ht]
  \centering
  \includegraphics[width=\textwidth]{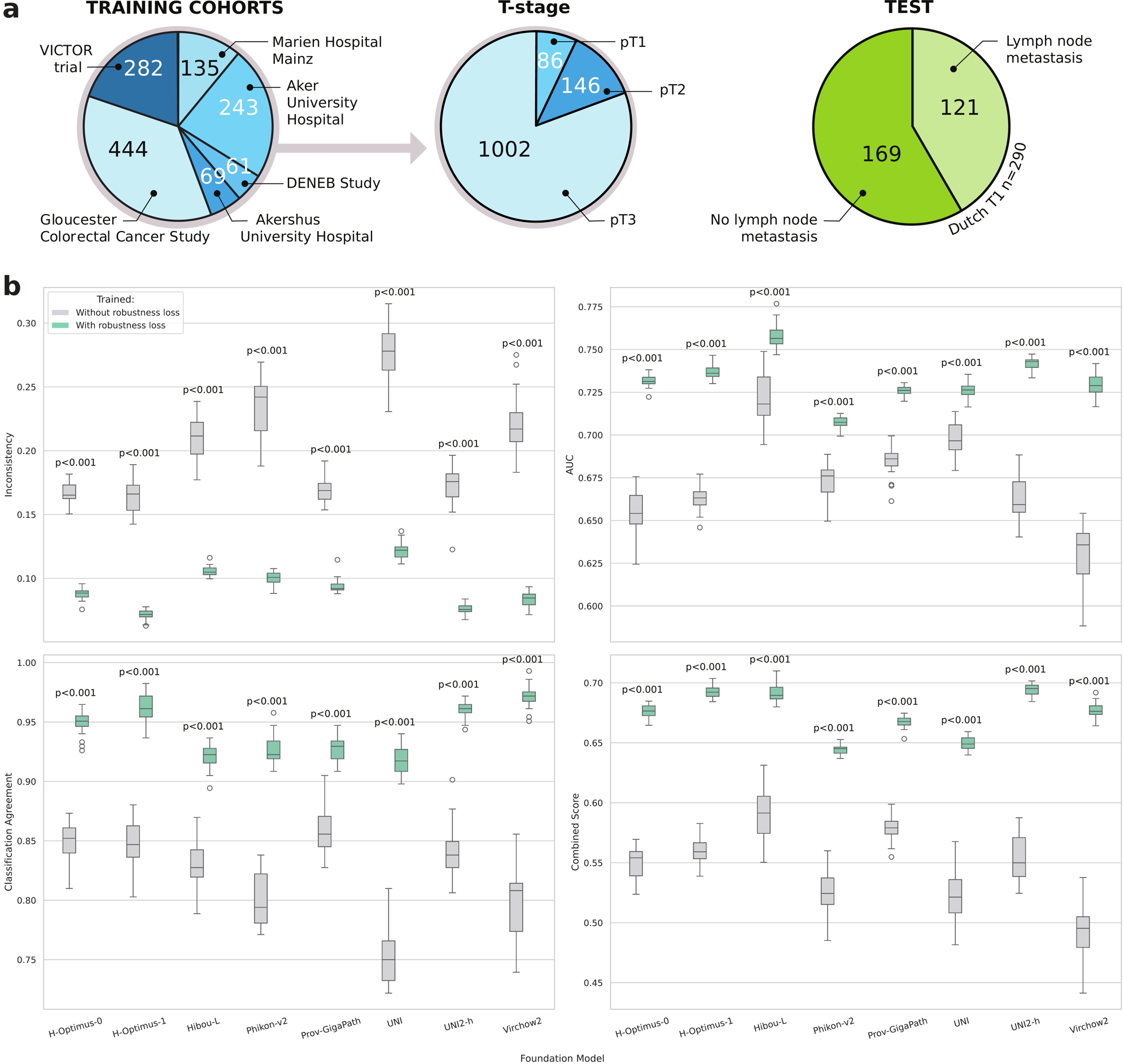}
  \caption[Prediction of lymph node metastasis]{%
    \textbf{Prediction of lymph node metastasis}\\
  }\label{fig:lnm-comparison-boxplot}
\end{figure}

In order to demonstrate that our findings are not specific to a particular task such as survival prediction,
we repeat the experiments for prediction of LNM in pT1 CRC (\cref{fig:lnm-comparison-boxplot}). Similar improvements in robustness are observed across all foundation models when downstream task-specific models are trained with robustness losses (all p < 0.001), with inconsistency decreasing by 79\% to 163\% and classification agreement increasing by 96\% to 593\% (Extended Data \cref{e-tab:test-lnm-improvement}). In this prediction task, the improvement in prediction accuracy by training with robustness loss is substantial and statistically significant (p < 0.001) for all foundation models. While only the Hibou-L features gave an average Area Under the receiver operating characteristic Curve (AUC) of above 0.7 when training without robustness loss, the average AUC is above 0.7 for all
foundation models when training with robustness loss.
For Virchow2, we see the largest change in average AUC, improving from 0.63 without robustness loss
to 0.73 with robustness loss (\cref{fig:lnm-comparison-boxplot}b), at the same time as the inconsistency decreases from 0.22 to 0.08 and the classification agreement increases from 80\% to 97\% (Extended Data \cref{e-tab:test-lnm-improvement}).

Interestingly, the relative performance of the foundation models differs between the
two prediction tasks, both in terms of robustness and prediction accuracy.
For instance, Hibou-L has the lowest robustness and the lowest prediction accuracy in the survival prediction task, yet achieves the highest prediction accuracy in the LNM prediction task. In the LNM prediction task, Hibou-L also performs comparably to the best-performing models in terms of robustness when trained without robustness loss and it improves markedly by training with robustness loss, although to a lesser extent than the other foundation models. By contrast, UNI performs well in the survival prediction task but shows poor robustness in the LNM prediction task, especially when trained without the robustness loss. Despite these task-specific differences in baseline performance, adding robustness loss substantially improves the robustness and generally also improves the prediction accuracy in both tasks.

\subsection{Balancing robustness and prediction accuracy}
The balance between robustness and prediction accuracy is controlled by the weight factor that scales the robustness losses.
Adding the robustness losses with a small weight around 1 markedly increases both robustness and prediction accuracy for all foundation models in both prediction tasks.
Further increasing the weight steadily improves robustness metrics, whereas excessively high weights reduce prediction accuracy (see Extended Data \cref{e-fig:metric-vs-weight_outcome,e-fig:metric-vs-weight_lnm} and Supplementary Information for detailed results).
We selected a weight for each foundation model by optimising a score combining inconsistency and c-index in the tuning dataset (see \emph{Methods} for more details). For this selection of weight, the downstream task-specific models are both substantially more robust and have similar or better prediction accuracy in terms of c-index in external test data (\cref{fig:outcome-comparison-boxplot}b). 

The optimal weight for the robustness losses in the survival prediction task is 40, 25, 40, 125, 150, 20, 30 and 100 for the foundation models Phikon-v2, Hibou-L, UNI, Prov-GigaPath, Virchow2, H-Optimus-0, H-Optimus-1 and UNI2, respectively. For the LNM prediction task, the corresponding optimal weights are 25, 7.5, 100, 40, 150, 75, 125 and 75, respectively. The point of optimal balance between robustness and prediction accuracy varies considerably between foundation models and between prediction tasks, suggesting that a tuning set is likely required to determine a suitable weight for the robustness losses in many settings.

\subsection{Impact of individual robustness losses}
We next assess the contribution of each robustness loss term by performing ablation experiments in which models were trained with the score loss alone or the embedding loss alone instead of both losses. 
For the survival prediction task (comprehensive results in Extended Data \cref{e-tab:test-outcome-improvement} and Extended Data \cref{e-fig:outcome_model_comparison_all_3}), the embedding loss alone provides modest and variable benefits, decreasing inconsistency in all foundation models except H-Optimus-1, with relative improvement ranging from 2\% to 105\% and improving CCC by 108\% to 282\%, while showing limited gains in c-index and inconsistent effects on classification agreement.
By contrast, the score loss alone improves robustness more substantially and systematically, reducing inconsistency across all foundation models, with relative improvements ranging from 168\% to 265\%, increasing classification agreement in all models by 54\% to 261\% and improving CCC by 231\% to 1,914\%. Gains in c-index are limited and inconsistent when either the embedding or score loss is used alone. 
A similar pattern of improvements in robustness metrics is observed for the LNM prediction task (Extended Data \cref{e-tab:test-lnm-improvement} and Extended Data \cref{e-fig:lnm_model_comparison_all_3}). For both tasks, models trained with the score loss alone perform similarly to those trained with both losses on robustness metrics, whereas the combined approach yields more consistent improvements in predictive performance. The effect of loss weighting on robustness and prediction accuracy metrics is also similar for the score loss only and the combined approach, whereas the trends are weaker and less consistent with embedding loss only (Supplementary Figs. 11-22). Together, these findings support the conclusion that the score loss accounts for most of the robustness benefit, whereas combining both loss terms yields the most consistent overall improvement.

\clearpage
\section{Discussion}

Much of the interest in foundation models for computational pathology comes from the expectation that exposure to large and diverse training datasets should produce representations that generalise across institutions and technical settings. Our results, together with growing evidence from others, challenge this assumption. \autocite{mulliqi2025,dejong2025, thiringer2026, komen2024do, komen2025towards} 
We show that non-biological information contained within foundation model features is not only consistently represented across datasets but typically constitutes the most characteristic attribute of these representations.
Using features from foundation models to train deep learning networks for survival prediction
and LNM prediction, we demonstrate that the resulting models produce substantially different predictions for the same tissue section imaged with different scanners, as well as for tissue sections from the same tumour 
prepared at different sites.
We show that this variability can be effectively mitigated by introducing additional terms into the training loss of the downstream model, without altering either the underlying foundation models or the downstream model architecture. Evaluated using several robustness metrics, this approach greatly improves robustness across all eight widely used foundation models in computational pathology on large external datasets, regardless of baseline performance.
These external datasets include additional tissue sections prepared in different laboratories and WSIs acquired using different scanners than those used in the training.

An important finding of this study is that an appropriate balance between robustness and predictive performance can be achieved, at which substantial gains in robustness are accompanied by significant improvements in prediction accuracy for most foundation models and tasks. A possible explanation for the improved accuracy is that the regularisation imposed by the robustness losses may guide the optimisation towards solutions that rely more strongly on the underlying tissue biology, thereby improving the correctness of the predictions. This regularisation may also account for the less extreme prediction scores observed in models incorporating the robustness loss compared with those trained without it. Mulliqi et al. reported that models based on UNI and Virchow2 representations showed no intrinsic generalisation advantage over fully end-to-end task-specific models for prostate cancer diagnosis and Gleason grading.\autocite{mulliqi2025} This raises the possibility that technical variation in foundation model representations was not adequately accounted for during downstream training. However, the finding may also reflect the unusually large training dataset used in that study, comprising more than 100,000 biopsies from 7,342 patients.

Strong performance on research datasets alone is not sufficient for clinical implementation of deep learning models in computational pathology.\autocite{Kleppe2021DesigningDL,vanderLaak2021DeepLI} Variation in tissue processing, staining and scanning is inherent to routine pathology practice, both between sites and within the same site over time, and even subtle differences can introduce site-specific patterns that compromise generalisation when they correlate with the outcome in the training data.\autocite {howard2021, Hoque2023StainNM} As a result, good performance in one setting may partly reflect technical variation rather than underlying biology and fail to transfer to another setting. In our experiments, the consequences were substantial, with average classification agreement between WSIs of the same patient for the survival prediction task ranging from 75\% to 91\%. This is particularly concerning because survival prediction is a high-stakes task, where no directly interpretable ground truth is available at the time of use, and even small changes in performance may have meaningful clinical consequences.
Our findings therefore support treating robustness to technical variation as a prerequisite for clinical implementation.

While limitations in robustness may already be encoded in foundation model representations,\autocite{komen2025towards, Huang2025KnowledgeguidedAO, carloni2025} modifying these models directly is often impractical because retraining is costly and typically requires access to large pretraining datasets. Weng et al. further argued that foundation models should preserve as much information as possible, because information that appears irrelevant in one setting may still prove important in another.\autocite{Weng2024AnIA} Existing efforts have therefore largely focused on standardising the input image or reducing the influence of non-biological information on local image features. These include parameter-efficient adaptation of the feature extractor,\autocite{be2025lowrank} stain normalisation,\autocite{komen2025towards,mishra2025comparing,komen2024do,Huang2025KnowledgeguidedAO} and refinement of tile-level embeddings by aligning image features with domain-informed textual concepts.\autocite{Huang2025KnowledgeguidedAO} Kömen et al. \autocite{komen2025towards} explored approaches to mitigate technical variation between medical centres, including post-hoc correction of feature embeddings and domain-adversarial training of downstream models. However, the improvements reported so far have generally been limited. The approaches most similar to ours are those of Carloni et al. and Ryu et al., which used paired images of the same tissue section acquired with different scanners to enforce consistency.\autocite{carloni2025,Ryu2025SCORPIONAS} Carloni et al. applied this at the embedding level using a contrastive loss between scanner-paired WSIs, whereas Ryu et al. combined style-based augmentation with a prediction-level consistency loss for paired tiles in a tissue segmentation setting. Our results indicate that embedding loss alone produces only modest and inconsistent gains, in line with the limited robustness improvements reported by Carloni et al\autocite{carloni2025}. By contrast, directly regularising prediction scores produces substantially larger improvements and is essential for improving robustness of the final model output. This suggests that encouraging similarity in intermediate representations is not sufficient, because residual technical differences can still be amplified in later layers if they remain useful for reducing the task loss. Such behaviour parallels the well-recognised tendency of computational pathology models to rely on non-biological signals, even when biologically meaningful differences are more apparent to pathologists.\autocite{Stacke2020MeasuringDS} By explicitly constraining prediction scores, our approach reduces the opportunity for these residual technical signals to influence the final output. Notably, although the score loss accounts for most of the robustness gain, the combination of embedding and score losses produces the most consistent overall improvement across robustness and predictive performance.

Limitations of our study include the use of a single approach for pooling and classifying WSIs based on foundation model features, albeit the attention-based multiple instance learning approach has become very common in computational pathology for WSIs classification using foundation models.\autocite{Lu2024AVF, Xu2025WhenMI, Waqas2026TheNL}
Studies introducing foundation models often evaluate them on a large array of different
downstream tasks,\autocite{chen2024towards, Vorontsov2024AFM,zimmermann2024virchow2,filiot2024phikon} and it could be argued that limiting the evaluation to two tasks within the
same cancer type could be considered too narrow. Nonetheless, the robustness issues we identify are  not specific to the particular prediction task, as demonstrated by the fact that the foundation model features themselves include consistent information about scanners. Their magnitude may, however, vary by task, since simpler prediction tasks typically have stronger true signal.\autocite{Neidlinger2025BenchmarkingFM}
The method we propose relies on WSIs generated from the same tissue slides using different scanners in training, which may be impractical and expensive. However, this is not required when applying the trained models.  Once trained, they can be used on WSIs from a single scanner, as in routine pathology practice. We further hypothesise that synthetically generating a variant of each WSI that uniquely represents a slide in training will be sufficient to increase robustness, although likely not to the same extent as leveraging the natural variation present in multiple real WSIs.

In conclusion, foundation models have the potential to lower the bar for building high-performing and generalisable deep learning systems in computational pathology, but their clinical value is limited by insufficient robustness to technical variation. We show that these limitations can be mitigated without retraining the foundation models themselves by regularising the downstream task-specific network using robustness loss terms. By encouraging similar prediction scores for WSIs from the same slide acquired with different scanners, our approach steers downstream models towards biologically meaningful information, resulting in greatly improved robustness and, in many cases, higher prediction accuracy. These findings help enable the clinical use of foundation model-based systems by improving their reliability.

\clearpage
\section{Methods}

\subsection{Materials}

Nine cohorts were included in this study for survival prediction in CRC and LNM prediction in pT1 CRC. Some cohorts contributed to both analyses, whereas others were used for only one task (see Figs.~\ref{fig:outcome-comparison-boxplot}a and \ref{fig:lnm-comparison-boxplot}a). The retrospective use of patient data from these cohorts was approved by the Regional Committees for Medical and Health Research Ethics in Norway (REC), with separate approvals for survival prediction in CRC (REC reference no. 747764) and LNM prediction in pT1 CRC (REC reference no. 808481).

For survival prediction, four independent cohorts (Ahus, Aker, Gloucester and VICTOR) were used for training as previously described.\autocite{skrede2020deep}
Only patients under 85 years with distinct good (over five years follow-up post surgery and no record of recurrence or cancer-specific death) or distinct poor outcomes (death from cancer between 30 days inclusive and 3 years exclusive post surgery) were included for training.
Slides from these cohorts were digitised using two different scanners. A fifth independent cohort (QUASAR 2), comprising WSIs from five different scanners, was used for model selection. External test was performed using a sixth independent cohort (TransSCOT), where each patient typically had six associated WSIs; five WSIs acquired by imaging the same slide using each of five different scanners and one WSI acquired by imaging a different slide. Patient and WSI counts are presented in Supplementary Table 1 and baseline characteristics in Extended Data \cref{e-tab:baseline-characteristics_outcome}.

For LNM prediction, the training set included six independent cohorts (Ahus, Aker, Gloucester, Mainz, DENEB and VICTOR) scanned with three different scanners. Although the primary objective was prediction in pT1 CRC, patients with pT2 and pT3 tumours were additionally included in the training to increase sample size. External test was performed in a seventh independent cohort (Dutch T1). Patient and WSI counts are presented in Supplementary Table 2 and baseline characteristics in Extended Data \cref{e-tab:baseline-characteristics_lnm}. 

As general exclusion criteria across cohorts, patients were excluded in cases of multiple primary colorectal tumors at diagnosis, prior colorectal cancer, pathological tumour stage outside pT1-3, missing or insufficient tumour tissue, or incomplete clinicopathological information.

Resected tumour specimens were formalin-fixed and paraffin-embedded (FFPE) according to standard clinical protocols. FFPE blocks were sectioned at 3 \textmu{}m, stained with H\&E and digitised at the highest resolution available (40$\times$ magnification) to generate WSIs. Scanning was performed using Aperio AT2 and GT 450 DX (Leica Biosystems, Germany); NanoZoomer XR, NanoZooomer S210, NanoZoomer 2.0-HT and NanoZoomer S60 (Hamamatsu Photonics, Japan); KF-PRO-400 (KFBIO, China); and Pannoramic 1000 (3DHISTECH, Hungary). Digital WSI files were read using the Python interface of the
OpenSlide C library version 3.4.1.\autocite{goode2013openslide} For the development cohorts except the DENEB cohort, inclusion required the availability of paired WSIs of the same slide generated using both the Aperio AT2 and NanoZoomer XR scanners. For the DENEB cohort, only one WSI was available for each slide and an artificial second WSI was created for each slide by applying heavy random distortion consistently to the original WSI.

Depending on the cohort, FFPE blocks were either sectioned, stained and scanned at the Institute for Cancer Genetics and Informatics (ICGI), Oslo University Hospital, Norway, or H\&E-stained slides or WSIs were provided directly by the contributing institution. The type of material received from each institution is specified below for each cohort separately.

\subsubsection{Ahus}

The Ahus cohort comprised a retrospective consecutive series of 224 patients who underwent surgical resection for colon cancer at Akershus University Hospital, Norway, between 1988 and 2000.\autocite{bondi2005expression,skrede2020deep} FFPE tissue blocks were transferred to ICGI, where one H\&E-stained slide per patient was prepared and digitised using both Aperio AT2 and NanoZoomer XR scanners. 
After application of the general exclusion criteria (Supplementary Fig. 1), 206 patients were eligible. Of these, 69 were included in the LNM prediction analysis and 58 in the survival outcome prediction analysis.

\subsubsection{Aker}

The Aker cohort consisted of a retrospective series of 1,214 patients who underwent surgical resection for colorectal cancer at Aker Hospital (now part of Oslo University Hospital), Norway, between 1993 and 2003.
In the present study, a subset of patients with stage I–III disease and available resected tissue sections, previously analyzed\autocite{skrede2020deep,merok2013microsatellite,hveem2014prognostic}, was included.
One H\&E-stained slide per patient was received at ICGI and digitised using Aperio AT2 and NanoZoomer XR scanners.
After application of the general exclusion criteria (Supplementary Fig. 2), 625 patients were eligible. Of these, 243 were included in the LNM prediction analysis and 388 in the survival outcome prediction analysis.

\subsubsection{Gloucester}

The Gloucester cohort comprised 1,050 patients recruited to the Gloucester Colorectal Cancer Study (Cheltenham, UK) between 1988 and 1996.\autocite{petersen2002identification,skrede2020deep}
This prospective, consecutive cohort included patients who underwent surgical resection for colorectal cancer at Gloucestershire Royal Hospital during this period.
The study was designed to investigate the prognostic impact of established pathological factors on survival.
FFPE tissue blocks were transferred to ICGI, where one H\&E-stained slide per patient was prepared and digitised using both Aperio AT2 and NanoZoomer XR scanners.
After application of general exclusion criteria (Supplementary Fig. 3), 1,001 patients were eligible. Of these, 444 were included in the LNM prediction analysis and 265 in the survival outcome prediction analysis.

\subsubsection{VICTOR}

The VICTOR cohort included patients enrolled in the previously reported VICTOR randomised clinical trial (ISRCTN98278138) conducted in the United Kingdom between 2002 and 2004.\autocite{kerr2007rofecoxib, Midgley2010PhaseIR} The trial included 2,327 patients with histologically confirmed stage II or III colorectal cancer who had undergone surgical resection  and were randomised to receive adjuvant rofecoxib or placebo across 151 UK hospitals. Rofecoxib did not improve overall survival or reduce recurrence in the overall study population. For the present study, 795 patients were included for whom FFPE tissue blocks were collected as part of the translational study entitled “Investigating molecular markers in samples collected from patients who took part in the QUASAR 2, VICTOR and SCOT trials.”\autocite{Midgley2010PhaseIR} H\&E-stained slides prepared in Oxford, UK and were digitised at ICGI using both Aperio AT2 and NanoZoomer XR scanners. After application of exclusion criteria as desribed previously~\autocite{skrede2020deep}, 768 patients were eligible (Supplementary Fig. 4). Of these, 282 were included in the LNM prediction analysis and 427 in the survival outcome prediction analysis.

\subsubsection{QUASAR 2}
The QUASAR 2 cohort comprised 1,952 patients enrolled in the previously reported QUASAR~2 randomised clinical trial (ISRCTN45133151) conducted between 2005 and 2010.\autocite{kerr2016adjuvant} The trial included patients with histologically confirmed stage III or high-risk stage II colorectal cancer who had undergone surgical resection and were randomised to receive adjuvant capecitabine with or without bevacizumab across 170 hospitals in seven countries. The addition of bevacizumab did not improve outcomes in the overall study population. For the present study, 1,251 patients were included for whom FFPE tissue blocks were collected as part of the translational study entitled “Investigating molecular markers in samples collected from patients who took part in the QUASAR 2, VICTOR and SCOT trials.” 
One H\&E-stained slide per patient was prepared from FFPE blocks at ICGI. After application of exclusion criteria, 1,118 eligible patients were included (Supplementary Fig. 5).
Slides were scanned on five scanners, resulting in 1,110 Aperio AT2, 1,106 NanoZoomer XR, 1,105 KF-PRO-400, 1,084 Aperio GT450 DX, and 1,096 Pannoramic 1000 WSIs.

\subsubsection{TransSCOT}
The TransSCOT cohort comprised patients enrolled in the SCOT randomised clinical trial (ISRCTN59757862), conducted in the United Kingdom between 2008 and 2013.\autocite{Iveson20183V6}
The trial included 6,088 patients with histologically confirmed stage III or high-risk stage II colorectal cancer who had undergone surgical resection and were eligible for adjuvant oxaliplatin-based chemotherapy. Eligible patients were $\geq$18 years old, had WHO performance status 0-1, adequate organ function and no significant comorbidities limiting life expectancy. Participants were randomised to receive 3 or 6 months of adjuvant oxaliplatin-fluoropyrimidine therapy and 3 months of treatment was demonstrated to be non-inferior to 6 months.
For the present study, 3,182 patients from the translational arm of the SCOT trial (TransSCOT) were included. H\&E-stained slides from 3,126 eligible patients were digitised by ICGI using five scanners, which, after exclusions, resulting in 2,856 Aperio AT2, 2,856 NanoZoomer XR, 2,856 KF-PRO-400, 2,547 Aperio GT450 DX and 2,534 Pannoramic 1000 WSIs (Supplementary Fig. 6b). In addition, a separate set of 3,126 H\&E-stained slides was digitised in in Oxford, UK, at a central laboratory associated with the trial office using NanoZoomer 2.0-HT or NanoZoomer S60 scanners. After exclusions, 2,856 patients had two corresponding tissue sections available: one digitised by ICGI across the five scanners and a second one digitised independently in the UK (Supplementary Fig. 6c).

\subsubsection{Mainz}

The Mainz cohort comprised a retrospective series of 598 patients who underwent surgical resection for colorectal cancer at the Marien Hospital Mainz, Germany, between 2002 and 2016 and were diagnosed at the Institute of Pathology of the University Medical Center Mainz (UMC Mainz), Germany. For 182 patients with sufficient available tissue material, three or more representative tissue blocks per patient were selected to produce H\&E-stained slides at the University Medical Center Mainz. In total, 735 slides were digitised at ICGI using Aperio AT2 and NanoZoomer XR scanners. After application of exclusion criteria (Supplementary Fig. 7), 135 patients and 500 paired WSIs were included in the LNM prediction analysis. 

\subsubsection{DENEB}

The DENEB cohort comprised patients enrolled in the prospective DENEB study, part of the GALAXY study within the CIRCULATE-Japan project. DENEB is a nationwide registry study conducted in Japan to evaluate the association between circulating tumour DNA (ctDNA) status and pathological risk factors, particularly LNM, in patients with pT1 CRC following complete local resection. The study planned to recruit 200 patients between 2021 and 2023 who had undergone complete local resection and were scheduled for additional intestinal resection with lymph node dissection based on standard pathological risk criteria for LNM.\autocite{Miyo2021DENEBDO}
WSIs were generated at the National Cancer Center Hospital East (Japan) using a NanoZoomer S210 scanner. ICGI received 202 WSIs from 202 patients and after application of exclusion criteria (Supplementary Fig. 8), 61 WSIs from 61 patients were included in the LNM prediction analysis.

\subsubsection{Dutch T1}

The Dutch T1 cohort comprised patients identified by the Dutch T1 CRC Working Group from 21 hospitals (1 academic and 20 non-academic) in the Netherlands between 2000 and 2017 through the Netherlands Cancer Registry. Following comprehensive review of all electronic medical records, cases were included in the multicenter T1 CRC registration cohort when the pathology report confirmed tumour invasion through the muscularis mucosae and into, but not beyond, the submucosa. Exclusion criteria included hereditary CRC syndromes, synchronous CRC (defined as CRC diagnosed within the previous 5 years or concurrently elsewhere in the colorectum), non-adenocarcinoma histology, inflammatory bowel disease, neoadjuvant radiotherapy and missing pathology or endoscopy reports. Four subcohorts were defined according to morphology (non-pedunculated vs pedunculated T1 CRCs) and time period (2000–2014 vs 2014–2017). Non-pedunculated cohorts used a case-cohort design, whereas pedunculated cohorts followed a 1:3 matched case-control design, as described previously. \autocite{Haasnoot2020AssociationsON,Backes2018HistologicFA} Of the selected T1 CRCs from the four subcohorts, original diagnostic H\&E-stained slides were retrieved from the originating hospitals for centralised review. An expert pathologist (ML), blinded to clinical data and outcomes, confirmed the diagnosis and selected one representative slide per patient according to a predefined protocol prioritizing the invasive front. A total of 522 slides were digitised at ICGI using Aperio AT2 and NanoZoomer XR scanners. After exclusions, 290 patients were included in the LNM prediction analysis (Supplementary Fig. 9).

\subsection{Method overview}

The proposed method is a modification to the standard attention-based multiple instance
learning\autocite{ilse2018attention} trained on top of tile features extracted from foundation
models for computational pathology.\autocite{bommasani2021opportunities}
The method intends to improve the resulting model in terms of robustness to technical
variation such as differences in laboratory procedures and scanning equipment.

We introduce two additional losses, in addition to the standard
classification loss.
Each loss compares outputs at different points in the network (\cref{fig:method}). The comparison is done on the same physical area of a slide scanned with two different scanners and aims to minimise difference in output between the two WSIs. 
To enable this comparison, each pair of WSIs from the same slide was tiled such that the corresponding
tiles represented the same physical area.

The proposed method was applied to tile features extracted from 8 different foundation models for
computational pathology:
H-optimus-0\autocite{hoptimus0},
H-optimus-1\autocite{hoptimus1},
Hibou\nobreakdash-L\autocite{nechaev2024hibou},
Phikon\nobreakdash-v2\autocite{filiot2024phikon},
Prov\,GigaPath\autocite{xu2024whole},
UNI\autocite{chen2024towards},
UNI\,2\nobreakdash-H\autocite{chen2024towards}
Virchow\nobreakdash-v2\autocite{zimmermann2024virchow2}.

\subsection{Whole slide image registration}

To produce pairs of corresponding tiles from two different WSIs of the same tissue slide, we registered the WSI from NanoZoomer XR (moving image) to the WSI from Aperio AT2 (fixed image) with elastix.\autocite{klein2009elastix}
For each slide, the corresponding NanoZoomer XR and Aperio AT2 WSIs were downsampled by a factor of 16, before being written as greyscale images.
Multi-resolution registration with 8 different levels was done using the EulerTransform followed by an AffineTransform.
The AdvancedNormalizedCorrelation metric in conjunction with the AdaptiveStochasticGradientDescent optimiser were used in both cases.

Ten areas within the tissue of the fixed image were sampled randomly in order to assess the quality of the registration.
The transformation was applied to the coordinates to the moving image and for each of the ten regions, a 512$\times$512 tile was extracted at full resolution for both the moving and fixed image.
The average registration correlation metric between the 10 tile pairs were used as the final metric.
Only WSIs with a lower correlation metric score than -0.85 were considered successful (note that this correlation metric ranges from 1 to -1, where -1 is perfect correlation).

\subsection{Tumour segmentation}

To label WSI regions as tumour or non-tumour, we used our previously published automatic tumour segmentation method.\autocite{skrede2025generalisation}
For the outcome prediction datasets, we applied minor updates to the method pre-processing described in the published segmentation method study, which we expected to have only a small effect on the segmentation results:
\begin{itemize}
    \item Use rounding instead of flooring when casting float to integer types.
    \item Upgrade Ubuntu version in docker image from 18.04 to 22.04 in order to upgrade pixman library from version 0.34 to 0.40.
    \item In the original method, the whole WSI was read at the appropriate level with OpenSlide and downsampled to 1 MPP (\textmu{}m/pixel), before overlapping tiles were cut from this image. In the updated version, overlapping tiles are read directly from the WSI with OpenSlide at the appropriate level, before they are downsampled to 1 MPP.
    \item When generating the 5 MPP overview image of the full scan, we use the image size provided by OpenSlide for the corresponding read level instead of computing it based on the downsampling factor at that level.
\end{itemize}
In addition, we applied a different post-processing method to the 8-bit valued score images produced by the neural network:
\begin{itemize}
    \item Change upper hysteresis threshold from 229 to 127.
    \item In pruning after thresholding, only remove the extra area generated by the lower hysteresis threshold if the region activates the pruning condition.
    \item Change the threshold in the pruning condition from 229 to 127 to reflect the corresponding change in hysteresis thresholding.
\end{itemize}

\subsection{Tiling}

WSIs are partitioned into tiles of size 224$\times$224 pixels at spatial resolution of 0.5 MPP (about 20$\times$ magnification).
WSIs are tiled without overlap from the top left and only tiles where at least 50\% of the pixels are labelled as tumour by the automatic segmentation method are included.
The size and spatial resolution were selected because seven of the eight foundation models included in this study were trained using this configuration.
The UNI model was trained on 256$\times$256 sized tiles, but accepts 224$\times$224 sized tiles.

For training, we used pairs of corresponding tiles extracted from WSIs generated using Aperio AT2 and NanoZoomer XR.\
Tile coordinates from the Aperio AT2 WSI were transformed to coordinates in the
NanoZoomer XR WSI using the transform from the registration of these two WSIs.
For the survival prediction in CRC, we included only slides with two WSIs and  successful registration.
For the LNM prediction in pT1 CRC, when only one WSI was available for a slide, tile pairs were constructed using one tile from
the available WSI and one distorted version of that tile (see \emph{Augmentation} for augmentation variants and magnitudes).
If the registration failed, we used tiles from the moving WSI together with corresponding distorted tiles. 
For slides in which part of the tissue was out of bounds in one of the WSIs, meaning that some tissue lay outside of the scanned area, we retained only tiles with corresponding physical regions in both WSIs.

\subsection{Feature extraction}
For each of the eight foundation models, a feature vector was extracted for every tile in every scan. Feature vectors were saved to disk in separate H5 files for each scan. To speed up feature extraction, features were extracted using automatic mixed precision and saved in half precision to reduce file size and increase read speed. Before each tile was passed through the foundation model, it was divided by 255 in order to convert it to a float in the [0, 1] range. It was standardised using the mean and standard deviation applied during the training of the foundation model.

During feature extraction, augmentation was applied to each tile. Augmentation parameters were sampled on a per-WSI basis, ensuring that all tiles from the same WSI were augmented identically. Features were only extracted for one augmentation configuration, meaning that the augmentation was fixed for the entire training run. %

For survival prediction in CRC, all WSIs were augmented. For LNM prediction in pT1 CRC, an unaugmented feature vector was extracted for each scan. For WSIs without a corresponding image from a different scanner, an augmented version was generated to serve as a surrogate for that scanner. 

For augmented WSIs, the specific augmentation was sampled randomly during feature extraction, so the same WSI was not augmented consistently across foundation models.

\subsubsection{Augmentation}
For each scan, the following augmentation parameters were sampled at random.

\begin{equation*}
\begin{aligned}
\Delta h & \sim U(-0.2,\,0.2) \\
\alpha_s & \sim U\!\left(\tfrac{1}{3},\,3\right) \\
\alpha_v & \sim U(0.5, \,2) \\
\alpha_c & \sim U(0.5, \,2) \\
\eta_{p} & \sim U(0, \,0.02)\quad\text{independently for each pixel }p \\
\end{aligned}
\end{equation*}

where $\Delta h$ is the hue shift, $\alpha_s$ is the saturation scaling factor, $\alpha_v$ is the value (brightness) scaling factor, $\alpha_c$ is the contrast adjustment factor, and $\eta_p$ is the per-pixel additive noise term.

The image was first transformed to the hsv space. The $\Delta h$ was added to the hue channel, with a floating point modulo in order to make the hue channel stay in the [0, 1] range. The saturation channel was scaled by $\alpha_s$, before scaling the value channel by $\alpha_v$. Next, the image was clipped to [0, 1], before being converted back to the RGB colour space. Contrast was then adjusted with $\alpha_c$ using the adjust\_contrast function in PyTorch. Noise was added on a per pixel basis using $\eta_{p}$, before finally clipping it to the [0, 1] range again.

\subsection{Network architecture}

The complete network consists of a fixed network (encoder) unique to each foundation
model and a classification network (head) that we train using attention-based multiple
instance learning.\autocite{ilse2018attention}

During training, the network takes as input a batch of randomly sampled bags of 512 feature vectors extracted from a foundation model. The size of the feature vectors depends on the foundation model used. The feature vectors are then passed through three linear layers, each with 256 output neurons, to produce the final tile-level feature vector. The three linear layers, as opposed to the standard one layer, is to ensure the model has the expressivity needed to make the feature vectors similar across scanners. After each linear layer is a leaky ReLU activation function with a negative slope of 0.01, following by a 1-dimensional batch norm.\autocite{maas2013rectifier,ioffe2015batch}

For each of these final tile-level feature vectors, an attention score is calculated by passing it through a linear layer with 128 output neurons, followed by a hyperbolic tangent activation function and then another linear layer with 1 output neuron. The attention score is then calculated by taking the softmax of this value over all attention values in the bag.

A scan-level feature vector is then calculated as the weighted average of all tile-level feature vectors in the bag, using their attention score as weights. A 1-dimensional batch norm is subsequently applied, followed by a dropout with a rate of 0.2.

Finally, for each of the two tasks in this study, the network generates a prediction through a linear layer with two output neurons, corresponding to
the two task-specific classes. 
These two unbounded values are the final outputs of the network.

\subsection{Loss terms}

We introduce two additional loss terms during training, intended to increase the robustness of the trained model to scanner differences. Their contributions are controlled by scalar weights relative to the classification loss. We set the two weights equal and sweep a shared value $\lambda\in[0, 1000]$.

The first additional loss function is the embedding loss. We use a modified InfoNCE as a contrastive loss on the tile-level embeddings taken from the final layer of the encoder\autocite{oord2018representation}. For each tile-level embedding in the bag, a tile-level embedding corresponding to the exact same physical area scanned on a different scanner were used as the positive sample. Then, five randomly sampled tile-level embeddings from a different patient, but the same scanner, were used as negative samples. 

\newcommand{\simfn}{\operatorname{sim}}

For each tile index $i\in\{1,\dots,N\}$, let $\mathbf{z}_i^{a}$ and $\mathbf{z}_i^{b}$ denote the corresponding $\ell_2$-normalised embedding vectors from scanners $a$ and $b$.
We use cosine similarity $\simfn(\mathbf{u},\mathbf{v})=\mathbf{u}^\top\mathbf{v}$ and temperature $\tau=0.1$.
We treat $\mathbf{z}_i^{a}$ as the anchor and $\mathbf{z}_i^{b}$ as the matched positive for tile $i$.
Let $\mathcal{N}_i^{a}$ be a set of indices for five negative tiles sampled for tile $i$ from a different patient scanned on scanner $a$.
We define
\begin{equation*}
\ell_i^{a\to b}
=
-\log
\frac{\exp\left(\simfn(\mathbf{z}_i^{a},\mathbf{z}_i^{b})/\tau\right)}
{\exp\left(\simfn(\mathbf{z}_i^{a},\mathbf{z}_i^{b})/\tau\right)
+\sum_{j\in\mathcal{N}_i^{a}}\exp\left(\simfn(\mathbf{z}_i^{a},\mathbf{z}_j^{a})/\tau\right)}
\end{equation*}
and average it over all positive tiles
\begin{equation*}
\mathcal{L}_{a\to b}=\frac{1}{N}\sum_{i=1}^N \ell_i^{a\to b}.
\end{equation*}

The embedding loss is then given as

\begin{equation*}
\mathcal{L}_{\mathrm{Embedding}}
=
\frac{1}{2}\left(
  \mathcal{L}_{\mathrm{AT2}\to\mathrm{XR}}
  +
  \mathcal{L}_{\mathrm{XR}\to\mathrm{AT2}}
\right).
\end{equation*}

The second additional loss is the score loss, defined as the mean-squared error on the score prediction between two
registered WSIs from different scanners, intended to force the model to give the same
prediction regardless of scanner. More formally, for each paired WSI $i\in\{1,\dots,N\}$ let $\widehat{\mathbf{s}}_i^{a}\in\mathbb{R}^{C}$ and $\widehat{\mathbf{s}}_i^{b}\in\mathbb{R}^{C}$ denote the model score vectors (e.g., logits) predicted from scanners $a$ and $b$, respectively, where $C$ is the number of classes to predict over. We then define the score loss as
\begin{equation*}
  \mathcal{L}_{\mathrm{Score}}
  =
  \frac{1}{NC}\sum_{i=1}^{N}\sum_{c=1}^{C}\left(\widehat{s}_{i,c}^{\mathrm{AT2}}-\widehat{s}_{i,c}^{\mathrm{XR}}\right)^2.
\end{equation*}

The classification loss is a standard cross-entropy loss.
Let $\widehat{\mathbf{y}}_i\in\mathbb{R}^{C}$ denote the predicted logits for sample $i$ and $c_i\in\{1,\dots,C\}$ the corresponding ground-truth class index. The cross-entropy loss is then given by
\begin{equation*}
  \mathcal{L}_{\mathrm{Classification}}
  =
  \frac{1}{N}\sum_{i=1}^{N}
  \left(
    -\log
    \frac{\exp(\widehat{y}_{i,c_i})}{\sum_{j=1}^{C}\exp(\widehat{y}_{i,j})}
  \right).
\end{equation*}

The total loss is then defined as 
\begin{equation}\label{eq:total-loss}
\mathcal{L}
=
\mathcal{L}_{\mathrm{Classification}}
+\lambda\mathcal{L}_{\mathrm{Embedding}}
+\lambda\mathcal{L}_{\mathrm{Score}}.
\end{equation}
where $\lambda$ is the weighting of the embedding and score terms.

\subsection{Dataset balancing}
For LNM prediction in pT1 CRC, the slides were oversampled with respect to T stage because most samples originated from pT2 or pT3 tumours. 
For outcome prediction in CRC, the slides were oversampled based on outcome to obtain equal numbers from each class. In both tasks, the minority groups were oversampled at the start of each epoch until they contained the same number of samples as the majority group, thereby ensuring balanced class distributions.

\subsection{Training}

For each training step, a batch of size $B=16$ was sampled, where each of the $B$ samples
consisted of a pair of WSIs (or a single WSI and an augmented variant if no corresponding WSI was available).
From each WSI pair in the batch, a bag of 512 corresponding tiles was
then randomly selected without replacement.

Each individual model was trained using the hyperparameters detailed below, with a checkpoint saved every 500 model updates.
\begin{itemize}
  \item Number of model updates: 20,000 
  \item Batch size (WSI pairs): 16 
  \item Optimiser: SGD 
  \item Momentum: 0.9 
  \item Weight decay: 0.02 
  \item Learning rate scheduler: One Cycle LR 
  \item Maximum learning rate: 0.01 
  \item Gradient norm clipping: 20 
  \item Dropout: 0.5
  \item Cross entropy loss weight: 1 
  \item Mean squared error loss weight: 0--1,000 
  \item InfoNCE loss weight: 0--1000 
  \item InfoNCE negative samples: 5 
  \item InfoNCE temperature: 0.1 
\end{itemize}

For each of the eight foundation models, we trained models for each
of 28 loss weights $\lambda$ (from \cref{eq:total-loss}): [0, 0.01, 0.05, 0.1, 0.5, 1.0, 2.5, 5.0, 7.5, 10, 12.5, 15, 20, 25, 30, 40, 50, 75,
100, 125, 150, 200, 250, 400, 500, 600, 800, 1,000]. In order to calculate statistics for each training foundation model and weight combination, we ran 20 independent training runs for each combination.
A weight of 0 corresponds to a regular Attention-based multiple instance model trained
without additional losses. 

\subsection{Prediction score classification}

Each successfully analysed WSI is given a prediction score value between 0 and 1.
For CRC outcome prediction, the score reflects the risk of poor outcome and is
classified as \emph{Good outcome} if the score is $\leq$ 0.5 and as
\emph{Poor outcome} if the score is $>$ 0.5.
For CRC pT1 LNM prediction, the score reflects the risk of LNM and is
classified as \emph{LNM negative} if the score is $\leq$ 0.5 and as 
\emph{LNM positive} if the score is $>$ 0.5.

\subsection{Choosing the best loss term weight $\lambda$}
For each foundation model, we performed 20 training runs for each of the 28 loss weights $\lambda$ (from \cref{eq:total-loss}), selecting one checkpoint for each run during tuning. We then choose the best $\lambda$ as the one with the highest average combined score (see \cref{eq:selection}) across the 20 training runs. These weights are the ones visualised in \cref{fig:outcome-comparison-boxplot} and \cref{fig:lnm-comparison-boxplot}.

\subsection{Statistical analyses}

All tests for statistical significance were treated as two-sided tests against the null hypothesis.
Differences and/or effects in the compared groups were considered statistically significant
when P < 0.05.
Prognostic performance for the survival outcome prediction task was assessed using the Harrell's concordance index (c-index).
Classification performance for the T1 LNM prediction task was assessed using the area under the receiver operating characteristic curve (AUC).
In order to calculate the p-values displayed in \cref{fig:outcome-comparison-boxplot} and \cref{fig:lnm-comparison-boxplot}, we used the Mann–Whitney U test. In \cref{fig:cosine}, we used the Wilcoxon signed-rank test.  
Statistical analyses were done using Python 3.11.9. P-values were calculated using the Python package scipy version 1.15.1 and c-index was calculated using the Python package lifelines version 0.30.0  

\subsection{Inconsistency}

For a single model that can produce a score $y_{ps} \in [0, 1]$ for a patient $p
\in P$ and a scanner $s \in S$, we define
\begin{equation}
    inconsistency = \frac{\frac{1}{|P|} \sum_{p \in P} \sigma_p}{\sigma_{T}}
    \label{eq:robustness}
\end{equation}
where $P$ is the set of patients and $\sigma_p$ is the score standard deviation
for patient $p$ over all scanners $S$.
Similarly, $\sigma_{T}$ is the score standard deviation over all patients and
scanners in the QUASAR 2 dataset.

\subsection{Checkpoint selection}

For choosing the best model checkpoint from a training run in outcome prediction in CRC, we use a metric that balance prognostic ability and scanner
robustness
\begin{equation}
    c - k * inconsistency
    \label{eq:selection}
\end{equation}
where $c$ is Harrell's concordance index (c-index) over the model's patient
scores in the dataset and a patient score is obtained by averaging the
prediction score over all scanners.
Throughout, we use $k = 0.627$, which was found in internal experiments to balance the
5 to 95 percentile ranges of the $c-index$ values and the $k\times inconsistency$
values, when multiple different prediction models were run on the QUASAR 2 dataset.

For each training run, this metric was then used to choose which checkpoint to select for each training run. The checkpoints were chosen using the QUASAR 2 dataset. 

For LNM prediction in pT1 CRC, due to the lack of a tuning set, we always selected the last checkpoint during training.

\subsection{Classification agreement}

A single patient is classified according to the section \emph{Prediction score
classification} and can be classified differently by the same model depending on which
of the patient's WSIs are classified.
\emph{Classification agreement} measures the proportion of patients in a dataset that do
not change classification between two WSI variants (e.g. WSIs from two different scanners) averaged over all variant pairs.
That is, if $c(p, s)$ is the classification of a WSI from variant $s$ and patient $p$,
then the classification agreement is computed as
\begin{equation}
  \frac{1}{|Q|} \sum_{q \in Q} \frac{|\{p: c(p, s_i) = c(p, s_j)\}|}{|P|}
  \label{eq:classification_agreement}
\end{equation}
where $Q$ is the set of all variant pairs $q = (s_i, s_j)$ where 
$s_i \neq s_j$, and $P$ is the set of all patients.

\subsection{Robustness index}

Robustness index is designed to capture models that focus on biological information while ignoring confounding features such as different scanners and medical centers.\autocite{komen2025towards}
For each scan, it considers the 50 closest neighbours using cosine similarity.

The metric defines two categories. The first is Same biological class and Other confounding call (SO). In our case, these are WSIs with the same ground-truth label, but scanned on a different scanner. The second category is Other biological class and Same confounding class (OS). In our case, these are WSIs with a different ground-truth label, but scanned on the same scanner.

We consider the sizes of these subsets of the 50 nearest neighbors and define:

\begin{equation}
  RI = \frac{|SO|}{|SO| + |OS|}
  \label{eq:robustness_index}
\end{equation}

Robustness index is defined on the interval [0, 1]. In our case, RI = 0 indicates that scanner information dominates the embeddings and RI=1 means the biological information dominates the embeddings.

Note that we only consider patients with slides that were scanned on all five ICGI scanners, and we only considered one WSI per patient. This means that original scans were not included for this metric.

\subsection{Average rank to same patient}

Similarly to \cref{eq:robustness_index}, Average Rank to Same Patient (ARSP) is defined on embedding level. For each scan, all other WSIs are ranked by sorting in decreasing order of cosine similarity to $s$ in the embedding space (rank $1$ being the most similar). The ARSP is then defined as the average rank of the four other WSIs of the same slide as the WSI we are considering. Note that we only consider patients with slides that were scanned on all five ICGI scanners, and we only considered one WSI per patient. This means that original scans were not included for this metric.

More formally, we define
\begin{equation}
    \text{ARSP} = \frac{1}{|\mathcal{S}|} \sum_{s \in \mathcal{S}} \frac{1}{|R(s)|} \sum_{s' \in R(s)} \text{rank}(s', s),
    \label{eq:arsp}
\end{equation}

where $\mathcal{S}$ is the set of WSIs, $R(s)$ is the set of other WSIs of the same slide as $s$, and $\text{rank}(s', s)$ is the rank of $s'$ among all other WSIs when sorted by cosine similarity to $s$ (rank 1 being the most similar).

Thus, a model that completely ignores non-biological information would get a score of $(1+2+3+4)/4 = 2.5$

\subsection{Concordance correlation coefficient}

The concordance correlation coefficient (CCC), as defined by Lin et al. \autocite{lin1989}, measures the agreement between two sets of paired measurements. 

For a pair of scanners $(s_i, s_j)$, let $x_p$ and $y_p$ be the prediction scores for patient $p \in P$ on scanners $s_i$ and $s_j$. The CCC for this pair is defined as
\begin{equation}
    \text{CCC}(s_i, s_j) = \frac{2 \, \text{Cov}(x, y)}{\sigma_x^2 + \sigma_y^2 + (\mu_x - \mu_y)^2}
    \label{eq:ccc}
\end{equation}
where $\mu_x$ and $\sigma_x^2$ are the mean and variance of the prediction scores on scanner $s_i$ over all patients and $\mu_y$ and $\sigma_y^2$, being the same for $s_j$. $\text{Cov}(x, y)$ is the covariance between the paired scores.

We report the average CCC over all scanner pairs:
\begin{equation}
    \overline{\text{CCC}} = \frac{1}{|Q|} \sum_{(s_i, s_j) \in Q} \text{CCC}(s_i, s_j)
    \label{eq:ccc_avg}
\end{equation}
where $Q$ is the set of all scanner pairs $(s_i, s_j)$ with $s_i \neq s_j$.

CCC is defined on the interval $[-1, 1]$, where $\text{CCC} = 1$ indicates perfect agreement between the scanners, $\text{CCC} = 0$ indicates no agreement and $\text{CCC} = -1$ indicates perfect disagreement.

\subsection{Rank-based hazard ratio}

To reduce sensitivity to the absolute scale of prediction scores and to handle ties consistently, we compute the hazard ratio on ranked scores rather than raw scores. For each patient $p \in P$, we define the normalised rank
\begin{equation}
    r_p = \frac{\text{rank}(y_p)}{|P|} \in (0, 1],
    \label{eq:normalized_rank}
\end{equation}
where $y_p$ is the prediction score, the lowest score has rank $1$, and ties are broken randomly. We then fit a Cox proportional hazards model with $r_p$ as the sole covariate and report
\begin{equation}
    \text{HR}_{\text{rank}} = \exp(\beta),
    \label{eq:rank_hr}
\end{equation}
where $\beta$ is the estimated rank coefficient. $\text{HR}_{\text{rank}}$ then corresponds to the hazard ratio between a patient at the top of the ranking and one at the bottom.

\clearpage
\subsection{Author contributions}

LVDS, ED, JH, RWW, JE, AN, NAS, IT, DCW, RSK, YN, MM, SF, DNC, MML, and DJK provided access to samples, clinical data and pathological data.
LVDS, IK, JAN, HAA, DCW, YN, MM, SF and MML were responsible for sample preparation and imaging.
O-JS, LVDS, KC, WK, JK, MP, MXI, MML and AK decided which samples to include.
ALH, O-JS, SDR, and JK processed the digital images and performed the machine learning.
ALH, O-JS, KL, and AK the did the statistical analyses.
ALH, O-JS, LVDS, MP, JAN, KSDG, MC, TSH, KL, MN, MML, and AK interpreted the data and analyses.
ALH, O-JS, LVDS, KC, KL, SDR, JK and AK wrote the first draft of the manuscript, and all authors reviewed, contributed to, and approved the manuscript.
AK had the final responsibility for the decision to submit for publication.

\subsection{Declaration of interests}

O-JS, SDR, WK, MP, JAN, HAA, MXI, TSH, KL, MN, DJK, and AK report having shares in DoMore Diagnostics.
KL reports being a board member in DoMore Diagnostics.
ALH and SDR report being employed by DoMore Diagnostics.
O-JS, KL, and TSH report filing a patent application titled ``Histological image analysis'' with International Patent Number PCT/EP2018/080828.
O-JS, KL, TSH, and AK report filing a patent application titled ``Histological image analysis'' with International Patent Application Number PCT/EP2020/076090.

\subsection{Code availability}

The source code is made available to reviewers as a submitted zip archive file and will be made publicly available through GitHub upon publication of the paper.

\subsection{Data availability}

Individual patient-level data can be made available to other researchers upon reasonable
request by contacting the corresponding author, subject to approval by the relevant
people or review board at the institutions that provided the original data.

\subsection{Funding}

This study was funded by The Research Council of Norway (grant numbers 309610, 334862, and 357305). The SCOT trial was supported by the Medical Research Council (transferred to NETSCC—Efficacy and Mechanism Evaluation; grant reference G0601705), the Swedish Cancer Society, and Cancer Research UK Core Clinical Trials Unit Funding (Funding Ref: C6716/A9894); translational analyses were funded by The Oxford NIHR Comprehensive Biomedical Research Centre (BRC). The views expressed are those of the authors and not necessarily those of The Research Council of Norway, NHS, the NIHR, or the Department of Health. DNC is funded by a Cancer Research UK Senior Cancer Research Fellowship (RCCSCF-Nov24/100001).

\subsection{Acknowledgements}

We thank the laboratory and technical personnel at the Institute for Cancer Genetics and
Informatics for essential sample preparation and assistance;
Marian Seiergren for assisting with figures;
Ortomedic AS for providing the Aperio GT450DX scanner on a rental basis;
Akershus University hospital, Oslo University Hospital (Aker Hospital), Cheltenham
General Hospital, Marienhaus Hospital, National Hospital Organization Osaka National Hospital, Utrecht Medical Center,
for access to materials and the personnel at said institutions for sample preparation;
the participating centres in the SCOT, QUASAR 2 and the VICTOR trials; and all
participating patients.

\clearpage
\singlespacing%

\printbibliography%

\clearpage
 \section{Figure captions}

\textbf{\cref{fig:method} |}
\textbf{Method overview}
\textbf{a,} Conventional use of multiple-instance learning and foundational models in computational pathology for predicting attributes of a slide or a patient. The pipeline consists of partitioning the whole-slide image (WSI) into multiple image tiles, processing each image tile by the foundation model to produce features, applying a few fully-connected layers to produce tile features, and then pooling the tile features of different tiles of the WSI into a combined representation of the whole WSI, which are then used to predict the target outcome. \textbf{b,} In our novel approach, we propose to train using co-registered tiles from two different WSIs of the same slide. Robustness loss is added to penalise differences between the tile features of the co-registered tiles as well as differences between the prediction scores of the two WSIs. Note that analysis of new cases does not require multiple WSIs; the trained model is applied to a single WSI in precisely the same manner as for models trained using the conventional approach.

\textbf{\cref{fig:scan-comparison} |}
\textbf{Foundation model robustness issue}
\textbf{a,} Example of whole-slide image (WSI) of a tissue section. \textbf{b,} WSIs of a different section of the same tissue block as in \textbf{a}, acquired using five different scanners. \textbf{c,} The output from the foundation model UNI of 5 randomly selected tiles from all WSIs in the TransSCOT dataset were projected into two dimensions using t-SNE and then visualised using the origin of the tile as label. \textbf{d-e,} Scatter plot for all WSIs in the TransSCOT dataset showing the correlation between the prediction score for the WSI of an original slide against the prediction score for the corresponding new slide scanned on one of five scanners. The prediction scores are calculated from a model predicting survival of patients with early-stage colorectal cancer using features from the foundation model UNI. $r$ is the Pearson's correlation coefficient and $n$ is the number of WSIs. \textbf{d,} Prediction scores computed using a model trained without robustness loss. \textbf{e,} Prediction scores computed using a model trained with robustness loss, where the weight applied for the robustness loss was the one giving best average result in the QUASAR 2 tuning dataset when training 20 models for each weight.

\textbf{\cref{fig:outcome-comparison-boxplot} |}
\textbf{Survival prediction}
\textbf{a,} Datasets and number of patients used in this experiment. \textbf{b,} Comparison of robustness and performance in the TransSCOT dataset when predicting survival of patients with early-stage colorectal cancer using different foundation models with and without the robustness loss, where the weight applied for the robustness loss was the one giving best average combined score in tuning when training 20 models for each weight. The combined score is a weighted average of inconsistency and c-index.
Box plots mark the median, the 25th and 75th percentiles (IQR), 25th percentile - 1.5$\times$IQR, 75th percentile + 1.5$\times$IQR, and outliers of 20 models.
VICTOR, Vioxx in Colorectal cancer Therapy: definition of Optimal Regime;
QUASAR, QUick And Simple And Reliable;
SCOT, Short Course Oncology Therapy.

\textbf{\cref{fig:heatmap} |}
\textbf{Spatial robustness}
\textbf{a,} Difference from mean prediction score of same patient for each whole-slide image (WSI) in the TransSCOT dataset, sorted by the standard deviation of the prediction scores for the six WSIs of each patient. The prediction scores are calculated from a model predicting survival of patients with early-stage colorectal cancer using features from the foundation model UNI, trained with or without robustness loss. The weight applied for the robustness loss was the one giving best average result in the QUASAR 2 tuning dataset when training 20 models for each weight. \textbf{b,} The six WSIs of a representative patient and corresponding heatmaps showing the prediction score of the model trained with (middle row) or without (bottom row) robustness loss for each individual tile in the WSIs, superimposed on a greyscale version of the WSI. In the heatmaps, a green colour indicates low prediction scores and a red colour indicates high prediction scores.

\textbf{\cref{fig:cosine} |}
\textbf{Robustness in each layer}
For each layer after the foundation model, the features of each tile from all whole-slide images (WSIs) in the TransSCOT dataset were calculated from models trained to predict survival of patients with early-stage colorectal cancer. Average Rank to Same Patient and Robustness Index was calculated at each layer for all 20 models trained with and all 20 models trained without the robustness loss for each of the eight foundation models (H\nobreakdash-optimus\nobreakdash-0, H\nobreakdash-optimus\nobreakdash-1, Hibou\nobreakdash-L, Phikon\nobreakdash-v2, Prov\,GigaPath, UNI, UNI\,2\nobreakdash-H, and Virchow\nobreakdash-v2). The weight applied for the robustness loss was the one giving best result in the QUASAR 2 tuning dataset when choosing based on combined score. The upper box plot illustrates how tightly WSIs of the same patient are grouped, with a lower value indicating that the similarity between WSIs of one patient is higher than WSIs of different patients on the same scanner, thus the influence by non-biological differences smaller. The bottom box plot illustrates how much the model groups biological information over non-biological information, with a higher value indicating that WSIs are grouped more based on biology than scanner information, implying that the unique biological information is more distinctly and better characterised in the model features.
Box plots mark the median, the 25th and 75th percentiles (IQR), 25th percentile - 1.5$\times$IQR, 75th percentile + 1.5$\times$IQR, and outliers.  

\textbf{\cref{fig:lnm-comparison-boxplot} |}
\textbf{Prediction of lymph node metastasis}
\textbf{a,} Datasets and number of patients used in this experiment. \textbf{b,} Comparison of robustness and performance in the Dutch-T1 dataset when predicting lymph node metastasis in T1 colorectal cancer patients using different foundation models with and without the robustness loss, where the weight applied for the robustness loss was the one giving best average combined score in tuning when training 20 models for each weight. The combined score is a weighted average of inconsistency and c-index.
Box plots mark the median, the 25th and 75th percentiles (IQR), 25th percentile - 1.5$\times$IQR, 75th percentile + 1.5$\times$IQR, and outliers of 20 models.
VICTOR, Vioxx in Colorectal cancer Therapy: definition of Optimal Regime;
DENEB, Development of new criteria for curability after local excision of pathological T1 colorectal cancer using liquid biopsy;
AUC, Area Under the receiver operating characteristic Curve.

\clearpage
\section{Figures}

\clearpage
\section{Appendix}

\vspace{0.5cm}

The TransSCOT Trial Management Group includes (alphabetical order):

\vspace{0.2cm}

David Church\textsuperscript{1},
Enric Domingo\textsuperscript{2},
Joanne Edwards\textsuperscript{3},
Bengt Glimelius\textsuperscript{4},
Ismail Gogenur\textsuperscript{5},
Andrea Harkin\textsuperscript{6},
Jennifer Hay\textsuperscript{7},
Timothy Iveson\textsuperscript{8},
Emma Jaeger\textsuperscript{2},
Caroline Kelly\textsuperscript{6},
Rachel Kerr\textsuperscript{2},
Noori Maka\textsuperscript{7},
Karin Oien\textsuperscript{7},
Clare Orange\textsuperscript{9},
Claire Palles\textsuperscript{10},
Campbell Roxburgh\textsuperscript{3},
Owen Sansom\textsuperscript{11},
Mark Saunders\textsuperscript{12},
Ian Tomlinson\textsuperscript{2}.

\vspace{0.2cm}

\textsuperscript{1}Cancer Genomics and Immunology Group, The Wellcome Centre for Human Genetics, University of Oxford UK\\
\textsuperscript{2}Department of Oncology, University of Oxford, UK\\
\textsuperscript{3}School of Cancer Sciences, University of Glasgow, Glasgow, UK\\
\textsuperscript{4}Uppsala University, Uppsala, Sweden\\
\textsuperscript{5}Centre for Surgical Science, Zealand University Hospital, Denmark\\
\textsuperscript{6}CRUK Glasgow Clinical Trials Unit, University of Glasgow, Glasgow, UK\\
\textsuperscript{7}Glasgow Tissue Research Facility, University of Glasgow, Queen Elizabeth University Hospital, Glasgow, UK\\
\textsuperscript{8}University of Southampton, Southampton, UK\\
\textsuperscript{9}NHS Greater Glasgow and Clyde Biorepository, Glasgow, UK\\
\textsuperscript{10}University of Birmingham, Birmingham, UK\\
\textsuperscript{11}CRUK Beatson Institute of Cancer Research, Garscube Estate, Glasgow, UK\\
\textsuperscript{12}The Christie NHS Foundation Trust, Manchester, UK

\clearpage

\captionsetup[figure]{name=Extended Data Fig.,labelfont=bf}
\captionsetup[table]{name=Extended Data Table,labelfont=bf}
\setcounter{figure}{0}
\setcounter{table}{0}
\section{Extended data}

\newgeometry{left=1.0cm,right=1.0cm}
\clearpage
\begin{table}[ht]
  \centering
  \small
  \caption{%
    \textbf{Baseline characteristics for included cohorts in CRC outcome prediction.}
    Data are given as \emph{median (interquartile range)} or \emph{count (percentage)}.
    Time to \emph{event} statistics are based only on patients with the respective event.
    CSD: cancer-specific death.
  }\label{e-tab:baseline-characteristics_outcome}
  \begin{tabular}{lr@{\hskip 0.1cm}lr@{\hskip 0.1cm}lr@{\hskip 0.1cm}lr@{\hskip 0.1cm}lr@{\hskip 0.1cm}lr@{\hskip 0.1cm}lr@{\hskip 0.1cm}l}
\toprule
                              & \multicolumn{2}{c}{Ahus}  & \multicolumn{2}{c}{Aker}  & \multicolumn{2}{c}{Gloucester}  & \multicolumn{2}{c}{VICTOR}  & \multicolumn{2}{c}{Training}  & \multicolumn{2}{c}{QUASAR 2}  & \multicolumn{2}{c}{TransSCOT} \\
\midrule                                                                                                                                                                
Patient count                 &   58 &                   &  388 &                   &  265 &                   &  427 &                   & 1138 &                        & 1112 &                   & 2856 &                  \\
Age                           &      &                   &      &                   &      &                   &      &                   &      &                        &      &                   &      &                  \\
\quad Years                   &   71 &   (64--77)    &   70 &   (60--77)    &   68 &   (62--75)    &   64 &   (57--71)    &   67 &   (59--75)         &   65 &   (59--71)    &   65 &   (58--70)   \\
Sex                           &      &                   &      &                   &      &                   &      &                   &      &                        &      &                   &      &                  \\
\quad Female                  &   34 &    (59\%)         &  195 &    (50\%)         &  144 &    (54\%)         &  149 &    (35\%)         &  522 &    (46\%)              &  474 &    (43\%)         & 1126 &    (39\%)        \\
\quad Male                    &   24 &    (41\%)         &  193 &    (50\%)         &  121 &    (46\%)         &  278 &    (65\%)         &  616 &    (54\%)              &  638 &    (57\%)         & 1730 &    (61\%)        \\
CSD         &      &                   &      &                   &      &                   &      &                   &      &                        &      &                   &      &                  \\
\quad False                   &   43 &    (74\%)         &  287 &    (74\%)         &  178 &    (67\%)         &  371 &    (87\%)         &  879 &    (77\%)              &  955 &    (86\%)         & 2438 &    (85\%)        \\
\quad True                    &   15 &    (26\%)         &  101 &    (26\%)         &   87 &    (33\%)         &   56 &    (13\%)         &  259 &    (23\%)              &  157 &    (14\%)         &  387 &    (14\%)        \\
\quad Missing                 &    0 &                   &    0 &                   &    0 &                   &    0 &                   &    0 &                        &    0 &                   &   31 &     (1\%)        \\
Time to CSD &      &                   &      &                   &      &                   &      &                   &      &                        &      &                   &      &                  \\
\quad Years                   &  1.4 &  (0.8--2.2)    &  1.8 &  (1.2--2.5)    &  1.2 &  (0.8--1.8)    &  2.2 &  (1.6--2.6)    &  1.6 &  (1.1--2.3)         &  2.7 &  (1.7--3.6)    &  1.1 &  (0.8--1.8)   \\
Follow-up time                &      &                   &      &                   &      &                   &      &                   &      &                        &      &                   &      &                  \\
\quad Years                   &  5.7 &  (3.4--6.3)    &  7.7 &  (2.9--10.9)    &  5.6 &  (1.8--7.1)    &  5.7 &  (5.2--6.1)    &  5.9 &  (5.1--7.3)         &  4.7 &  (3.4--5.1)    &  6.0 &  (4.9--7.1)   \\
pT                            &      &                   &      &                   &      &                   &      &                   &      &                        &      &                   &      &                  \\
\quad pT1                     &    1 &     (2\%)         &   24 &     (6\%)         &    4 &     (2\%)         &    6 &     (1\%)         &   35 &     (3\%)              &   18 &     (2\%)         &   57 &     (2\%)        \\
\quad pT2                     &   11 &    (19\%)         &   79 &    (20\%)         &   22 &     (8\%)         &   33 &     (8\%)         &  145 &    (13\%)              &   70 &     (6\%)         &  237 &     (8\%)        \\
\quad pT3                     &   44 &    (76\%)         &  263 &    (68\%)         &  129 &    (49\%)         &  289 &    (68\%)         &  725 &    (64\%)              &  583 &    (52\%)         & 1690 &    (59\%)        \\
\quad pT4                     &    2 &     (3\%)         &   22 &     (6\%)         &  109 &    (41\%)         &   87 &    (20\%)         &  220 &    (19\%)              &  392 &    (35\%)         &  872 &    (31\%)        \\
\quad Missing                 &    0 &                   &    0 &                   &    1 &    (<1\%)         &   12 &     (3\%)         &   13 &     (1\%)              &   49 &     (4\%)         &    0 &                  \\
pN stage                      &      &                   &      &                   &      &                   &      &                   &      &                        &      &                   &      &                  \\
\quad pN0                     &   35 &    (60\%)         &  258 &    (66\%)         &  142 &    (54\%)         &  187 &    (44\%)         &  622 &    (55\%)              &  398 &    (36\%)         &  533 &    (19\%)        \\
\quad pN1                     &   15 &    (26\%)         &   99 &    (26\%)         &   63 &    (24\%)         &  161 &    (38\%)         &  338 &    (30\%)              &  507 &    (46\%)         & 1637 &    (57\%)        \\
\quad pN2                     &    8 &    (14\%)         &   30 &     (8\%)         &   60 &    (23\%)         &   67 &    (16\%)         &  165 &    (14\%)              &  182 &    (16\%)         &  686 &    (24\%)        \\
\quad Missing                 &    0 &                   &    1 &    (<1\%)         &    0 &                   &   12 &     (3\%)         &   13 &     (1\%)              &   25 &     (2\%)         &    0 &                  \\
\bottomrule
\end{tabular}

\end{table}

\clearpage
\begin{table}[ht]
  \centering
  \footnotesize
  \caption{%
    \textbf{Baseline characteristics for included cohorts in CRC T1 LNM prediction.}
    Data are given as \emph{median (interquartile range)} or \emph{count (percentage)}.
    Time to \emph{event} statistics are based only on patients with the respective event.
  }\label{e-tab:baseline-characteristics_lnm}
  \begin{tabular}{lr@{\hskip 0.1cm}lr@{\hskip 0.1cm}lr@{\hskip 0.1cm}lr@{\hskip 0.1cm}lr@{\hskip 0.1cm}lr@{\hskip 0.1cm}lr@{\hskip 0.1cm}lr@{\hskip 0.1cm}l}
\toprule
                  & \multicolumn{2}{c}{Aker} & \multicolumn{2}{c}{Ahus} & \multicolumn{2}{c}{Gloucester} & \multicolumn{2}{c}{VICTOR} & \multicolumn{2}{c}{Mainz} & \multicolumn{2}{c}{DENEB} & \multicolumn{2}{c}{Training} & \multicolumn{2}{c}{Dutch T1} \\
\midrule
Patient count     &   69 &           &  243 &              &  444 &               &  282 &               &  135 &               &   61 &            & 1234 &                &  290 &            \\
Age               &      &           &      &              &      &               &      &               &      &               &      &            &      &                &      &            \\
\quad Years       &   71 & (59--77)  &   73 &   (63--79)   &   70 &   (64--77)    &   64 &   (58--71)    &   73 &   (66--79)    &   61 &   (52--72) &   69 &   (61--77)     &   68 &   (63--74) \\
Sex               &      &           &      &              &      &               &      &               &      &               &      &            &      &                &      &            \\
\quad Female      &   36 &  (52\%)   &  128 &    (53\%)    &  182 &    (41\%)     &  102 &    (36\%)     &   60 &    (44\%)     &   31 &    (51\%)  &  539 &    (44\%)      &  126 &    (43\%)  \\
\quad Male        &   33 &  (48\%)   &  115 &    (47\%)    &  262 &    (59\%)     &  180 &    (64\%)     &   75 &    (56\%)     &   30 &    (49\%)  &  695 &    (56\%)      &  164 &    (57\%)  \\
pT stage          &      &           &      &              &      &               &      &               &      &               &      &            &      &                &      &            \\
\quad pT1         &    0 &           &    4 &     (2\%)    &    4 &     (1\%)     &    6 &     (2\%)     &   11 &     (8\%)     &   61 &   (100\%)  &   86 &     (7\%)      &  290 &   (100\%)  \\
\quad pT2         &    3 &   (4\%)   &   26 &    (11\%)    &   45 &    (10\%)     &   37 &    (13\%)     &   35 &    (26\%)     &    0 &            &  146 &    (12\%)      &    0 &            \\
\quad pT3         &   66 &  (96\%)   &  213 &    (88\%)    &  395 &    (89\%)     &  239 &    (85\%)     &   89 &    (66\%)     &    0 &            & 1002 &    (81\%)      &    0 &            \\
pN stage          &      &           &      &              &      &               &      &               &      &               &      &            &      &                &      &            \\
\quad pN0         &    0 &           &   80 &    (33\%)    &  246 &    (55\%)     &    0 &               &   91 &    (67\%)     &   47 &    (77\%)  &  464 &    (38\%)      &  169 &    (58\%)  \\
\quad pN1         &   56 &  (81\%)   &  131 &    (54\%)    &  130 &    (29\%)     &  200 &    (71\%)     &   34 &    (25\%)     &   12 &    (20\%)  &  563 &    (46\%)      &  110 &    (38\%)  \\
\quad pN2         &   13 &  (19\%)   &   32 &    (13\%)    &   68 &    (15\%)     &   82 &    (29\%)     &   10 &     (7\%)     &    2 &     (3\%)  &  207 &    (17\%)      &   11 &     (4\%)  \\
Examined lymph nodes &   &           &      &              &      &               &      &               &      &               &      &            &      &                &      &            \\
\quad 12 or fewer &    0 &           &   90 &    (37\%)    &   43 &    (10\%)     &    0 &               &    4 &     (3\%)     &    8 &    (13\%)  &  145 &    (12\%)      &   66 &    (23\%)  \\
\quad More than 12&    0 &           &  102 &    (42\%)    &  401 &    (90\%)     &    0 &               &  131 &    (97\%)     &   53 &    (87\%)  &  687 &    (56\%)      &  224 &    (77\%)  \\
\quad Missing     &   69 & (100\%)   &   51 &    (21\%)    &    0 &               &  282 &   (100\%)     &    0 &               &    0 &            &  402 &    (33\%)      &    0 &            \\
\bottomrule
\end{tabular}

\end{table}
\restoregeometry

\newgeometry{top=1.0cm,bottom=2.5cm}
\begin{figure}[ht]
  \centering
  \includegraphics[width=0.57\textwidth]{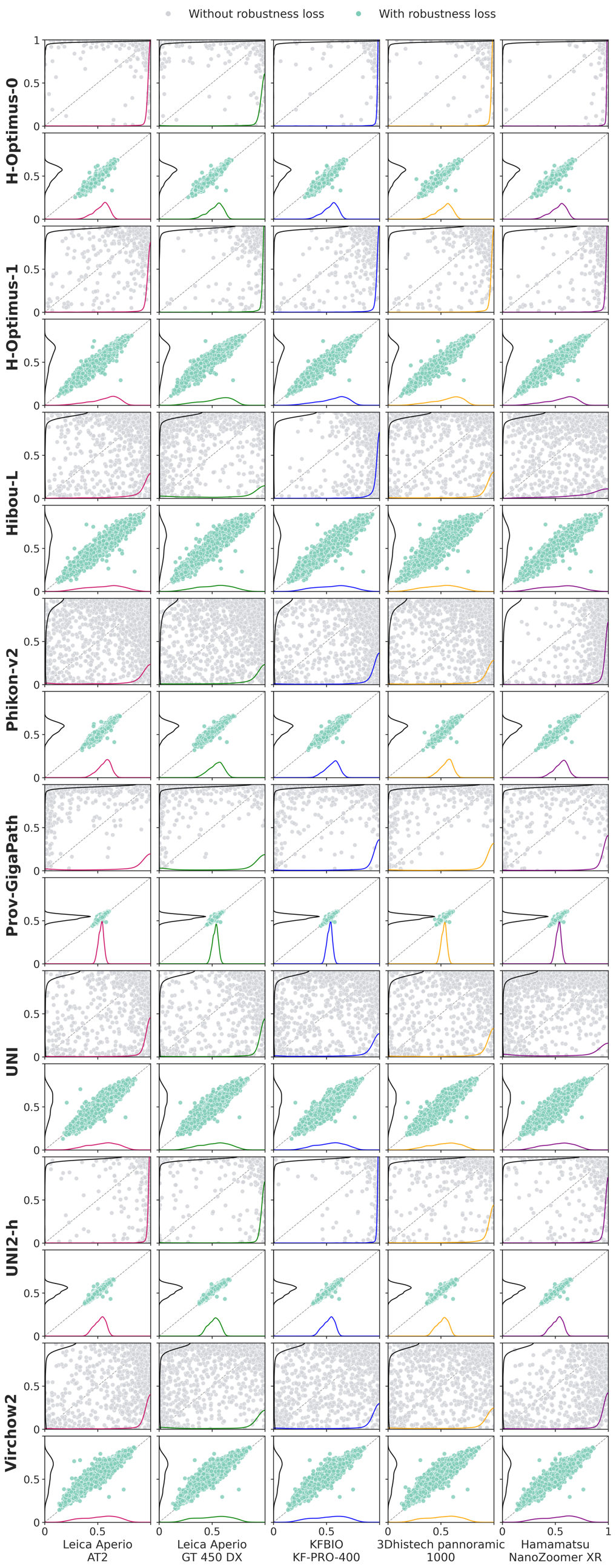}
  \caption{%
    \textbf{WSI prediction score comparison between tissue sections.}
    Comparison of model score between different tissue sections from the same patient. Each dot represents one tissue section. For each patient, the x-axis is the prediction score on a tissue section scanned at ICGI, on 5 different scanners, and the y-axis is the score of a original tissue section. For each scatterplot, kernel density estimates are shown for both the original tissue sections and the tissue sections scanned at ICGI.
  }\label{e-fig:scatterplots_KDE}
\end{figure}
\restoregeometry

\newgeometry{left=2cm,top=1.5cm,bottom=2.5cm}
\begin{table}[ht]
  \centering
  \small
  \setlength{\tabcolsep}{3pt}
  \caption{%
    \textbf{Performance comparison between different loss terms in survival prediction.}
    Metric summary results are given as \emph{mean}~$\pm$~\emph{standard deviation} (\emph{relative improvement}). For all columns with robustness loss, the best weight ($\lambda$) is used, and improvement is relative to no robustness loss ($\lambda$ = 0).
  }\label{e-tab:test-outcome-improvement}
  \begin{tabular}{llrr@{\hskip 0.1cm}rr@{\hskip 0.1cm}rr@{\hskip 0.1cm}r}
  \toprule
  & \textbf{Model} & \textbf{No robustness loss} & \multicolumn{2}{c}{\textbf{Embedding loss}} & \multicolumn{2}{c}{\textbf{Score loss}} & \multicolumn{2}{c}{\textbf{Both losses}} \\
  \midrule
  \multirow{8}{*}{\rotatebox[origin=c]{90}{\parbox{2cm}{\centering c-index}}}
   & H-optimus-0 & 0.661 $\pm$ 0.014 & 0.682 $\pm$ 0.013 & (6\%) & 0.689 $\pm$ 0.005 & (7\%) & 0.694 $\pm$ 0.006 & (10\%) \\
   & H-optimus-1 & 0.692 $\pm$ 0.011 & 0.689 $\pm$ 0.009 & (-2\%) & 0.695 $\pm$ 0.004 & (1\%) & 0.699 $\pm$ 0.005 & (2\%) \\
   & Hibou-L & 0.654 $\pm$ 0.012 & 0.678 $\pm$ 0.007 & (7\%) & 0.672 $\pm$ 0.005 & (6\%) & 0.680 $\pm$ 0.004 & (7\%) \\
   & Phikon-v2 & 0.654 $\pm$ 0.017 & 0.668 $\pm$ 0.014 & (4\%) & 0.666 $\pm$ 0.004 & (4\%) & 0.678 $\pm$ 0.006 & (7\%) \\
   & Prov-GigaPath & 0.662 $\pm$ 0.013 & 0.676 $\pm$ 0.008 & (1\%) & 0.673 $\pm$ 0.004 & (0\%) & 0.680 $\pm$ 0.007 & (5\%) \\
   & UNI & 0.681 $\pm$ 0.010 & 0.680 $\pm$ 0.012 & (0\%) & 0.670 $\pm$ 0.005 & (0\%) & 0.683 $\pm$ 0.005 & (0\%) \\
   & UNI2-H & 0.672 $\pm$ 0.015 & 0.690 $\pm$ 0.011 & (2\%) & 0.688 $\pm$ 0.005 & (2\%) & 0.697 $\pm$ 0.004 & (8\%) \\
   & Virchow2 & 0.680 $\pm$ 0.024 & 0.680 $\pm$ 0.011 & (1\%) & 0.662 $\pm$ 0.009 & (-2\%) & 0.682 $\pm$ 0.004 & (0\%) \\
  \midrule
  \multirow{8}{*}{\rotatebox[origin=c]{90}{\parbox{2cm}{\centering Rank-based Hazard Ratio}}} 
    & H-optimus-0 & 7.835 $\pm$ 1.675 & 10.648 $\pm$ 1.910 & (31\%) & 11.430 $\pm$ 0.930 & (37\%) & 12.427 $\pm$ 1.116 & (58\%) \\
    & H-optimus-1 & 12.294 $\pm$ 1.877 & 11.848 $\pm$ 1.867 & (-6\%) & 12.508 $\pm$ 0.874 & (2\%) & 13.483 $\pm$ 1.151 & (9\%) \\
    & Hibou-L & 6.932 $\pm$ 1.180 & 9.872 $\pm$ 1.205 & (36\%) & 8.765 $\pm$ 0.674 & (29\%) & 9.986 $\pm$ 0.739 & (44\%) \\
    & Phikon-v2 & 7.291 $\pm$ 1.774 & 8.719 $\pm$ 1.737 & (18\%) & 7.952 $\pm$ 0.468 & (15\%) & 9.847 $\pm$ 1.034 & (35\%) \\
    & Prov-GigaPath & 7.851 $\pm$ 1.547 & 9.597 $\pm$ 1.205 & (6\%) & 8.849 $\pm$ 0.498 & (-3\%) & 10.073 $\pm$ 1.131 & (28\%) \\
    & UNI & 10.216 $\pm$ 1.574 & 10.220 $\pm$ 1.759 & (2\%) & 8.568 $\pm$ 0.730 & (0\%) & 10.525 $\pm$ 0.845 & (3\%) \\
    & UNI2-H & 9.207 $\pm$ 1.967 & 11.985 $\pm$ 1.988 & (11\%) & 11.287 $\pm$ 0.866 & (8\%) & 13.036 $\pm$ 0.846 & (41\%) \\
    & Virchow2 & 10.720 $\pm$ 3.311 & 10.292 $\pm$ 1.555 & (7\%) & 7.655 $\pm$ 0.929 & (-9\%) & 10.322 $\pm$ 0.569 & (-3\%) \\
  \midrule
  \multirow{8}{*}{\rotatebox[origin=c]{90}{\parbox{2cm}{\centering Balanced accuracy}}}
   & H-optimus-0 & 0.547 $\pm$ 0.027 & 0.608 $\pm$ 0.027 & (18\%) & 0.634 $\pm$ 0.033 & (24\%) & 0.649 $\pm$ 0.011 & (29\%) \\
   & H-optimus-1 & 0.579 $\pm$ 0.039 & 0.632 $\pm$ 0.016 & (12\%) & 0.643 $\pm$ 0.012 & (14\%) & 0.652 $\pm$ 0.016 & (21\%) \\
   & Hibou-L & 0.570 $\pm$ 0.021 & 0.613 $\pm$ 0.023 & (10\%) & 0.600 $\pm$ 0.047 & (8\%) & 0.636 $\pm$ 0.013 & (18\%) \\
   & Phikon-v2 & 0.566 $\pm$ 0.026 & 0.606 $\pm$ 0.023 & (12\%) & 0.614 $\pm$ 0.024 & (16\%) & 0.628 $\pm$ 0.022 & (16\%) \\
   & Prov-GigaPath & 0.571 $\pm$ 0.024 & 0.621 $\pm$ 0.013 & (7\%) & 0.613 $\pm$ 0.021 & (7\%) & 0.596 $\pm$ 0.052 & (6\%) \\
   & UNI & 0.588 $\pm$ 0.031 & 0.619 $\pm$ 0.021 & (13\%) & 0.626 $\pm$ 0.027 & (16\%) & 0.636 $\pm$ 0.013 & (13\%) \\
   & UNI2-H & 0.551 $\pm$ 0.022 & 0.602 $\pm$ 0.028 & (7\%) & 0.644 $\pm$ 0.007 & (23\%) & 0.637 $\pm$ 0.032 & (23\%) \\
   & Virchow2 & 0.570 $\pm$ 0.039 & 0.612 $\pm$ 0.032 & (13\%) & 0.601 $\pm$ 0.035 & (9\%) & 0.620 $\pm$ 0.033 & (13\%) \\
  \midrule
  \multirow{8}{*}{\rotatebox[origin=c]{90}{\parbox{2cm}{\centering Combined score}}}
   & H-optimus-0 & 0.395 $\pm$ 0.053 & 0.488 $\pm$ 0.032 & (13\%) & 0.608 $\pm$ 0.012 & (50\%) & 0.591 $\pm$ 0.018 & (47\%) \\
   & H-optimus-1 & 0.498 $\pm$ 0.038 & 0.508 $\pm$ 0.040 & (-1\%) & 0.620 $\pm$ 0.007 & (33\%) & 0.609 $\pm$ 0.022 & (28\%) \\
   & Hibou-L & 0.248 $\pm$ 0.054 & 0.472 $\pm$ 0.031 & (45\%) & 0.526 $\pm$ 0.057 & (67\%) & 0.552 $\pm$ 0.026 & (67\%) \\
   & Phikon-v2 & 0.318 $\pm$ 0.064 & 0.414 $\pm$ 0.061 & (11\%) & 0.556 $\pm$ 0.011 & (47\%) & 0.535 $\pm$ 0.023 & (46\%) \\
   & Prov-GigaPath & 0.370 $\pm$ 0.059 & 0.442 $\pm$ 0.035 & (14\%) & 0.592 $\pm$ 0.006 & (51\%) & 0.578 $\pm$ 0.018 & (49\%) \\
   & UNI & 0.425 $\pm$ 0.040 & 0.495 $\pm$ 0.030 & (9\%) & 0.592 $\pm$ 0.011 & (43\%) & 0.590 $\pm$ 0.014 & (40\%) \\
   & UNI2-H & 0.401 $\pm$ 0.056 & 0.454 $\pm$ 0.046 & (9\%) & 0.605 $\pm$ 0.007 & (47\%) & 0.596 $\pm$ 0.021 & (48\%) \\
   & Virchow2 & 0.464 $\pm$ 0.045 & 0.518 $\pm$ 0.033 & (11\%) & 0.602 $\pm$ 0.010 & (39\%) & 0.602 $\pm$ 0.012 & (34\%) \\
  \midrule
  \multirow{8}{*}{\rotatebox[origin=c]{90}{\parbox{2cm}{\centering Inconsistency}}}
   & H-optimus-0 & 0.424 $\pm$ 0.082 & 0.309 $\pm$ 0.040 & (25\%) & 0.128 $\pm$ 0.022 & (216\%) & 0.164 $\pm$ 0.026 & (159\%) \\
   & H-optimus-1 & 0.310 $\pm$ 0.057 & 0.289 $\pm$ 0.061 & (2\%) & 0.119 $\pm$ 0.013 & (168\%) & 0.143 $\pm$ 0.032 & (116\%) \\
   & Hibou-L & 0.648 $\pm$ 0.086 & 0.330 $\pm$ 0.046 & (105\%) & 0.232 $\pm$ 0.094 & (204\%) & 0.204 $\pm$ 0.038 & (217\%) \\
   & Phikon-v2 & 0.536 $\pm$ 0.092 & 0.405 $\pm$ 0.084 & (20\%) & 0.175 $\pm$ 0.018 & (177\%) & 0.229 $\pm$ 0.031 & (134\%) \\
   & Prov-GigaPath & 0.466 $\pm$ 0.087 & 0.374 $\pm$ 0.053 & (33\%) & 0.129 $\pm$ 0.011 & (259\%) & 0.163 $\pm$ 0.025 & (186\%) \\
   & UNI & 0.408 $\pm$ 0.066 & 0.295 $\pm$ 0.040 & (24\%) & 0.125 $\pm$ 0.017 & (220\%) & 0.148 $\pm$ 0.021 & (175\%) \\
   & UNI2-H & 0.433 $\pm$ 0.086 & 0.376 $\pm$ 0.064 & (18\%) & 0.133 $\pm$ 0.015 & (215\%) & 0.161 $\pm$ 0.034 & (169\%) \\
   & Virchow2 & 0.344 $\pm$ 0.059 & 0.259 $\pm$ 0.045 & (31\%) & 0.097 $\pm$ 0.010 & (265\%) & 0.126 $\pm$ 0.020 & (172\%) \\
  \midrule
  \multirow{8}{*}{\rotatebox[origin=c]{90}{\parbox{2cm}{\centering Classification\\ agreement}}}
   & H-optimus-0 & 0.903 $\pm$ 0.059 & 0.902 $\pm$ 0.027 & (-29\%) & 0.945 $\pm$ 0.015 & (54\%) & 0.938 $\pm$ 0.015 & (56\%) \\
   & H-optimus-1 & 0.910 $\pm$ 0.053 & 0.883 $\pm$ 0.041 & (-27\%) & 0.947 $\pm$ 0.008 & (88\%) & 0.945 $\pm$ 0.016 & (65\%) \\
   & Hibou-L & 0.749 $\pm$ 0.089 & 0.890 $\pm$ 0.024 & (155\%) & 0.918 $\pm$ 0.043 & (261\%) & 0.920 $\pm$ 0.018 & (215\%) \\
   & Phikon-v2 & 0.820 $\pm$ 0.083 & 0.854 $\pm$ 0.038 & (-10\%) & 0.927 $\pm$ 0.017 & (58\%) & 0.907 $\pm$ 0.019 & (92\%) \\
   & Prov-GigaPath & 0.851 $\pm$ 0.062 & 0.832 $\pm$ 0.041 & (16\%) & 0.942 $\pm$ 0.011 & (180\%) & 0.953 $\pm$ 0.022 & (215\%) \\
   & UNI & 0.857 $\pm$ 0.055 & 0.887 $\pm$ 0.039 & (-10\%) & 0.949 $\pm$ 0.011 & (138\%) & 0.939 $\pm$ 0.011 & (136\%) \\
   & UNI2-H & 0.900 $\pm$ 0.044 & 0.869 $\pm$ 0.053 & (6\%) & 0.943 $\pm$ 0.009 & (81\%) & 0.934 $\pm$ 0.017 & (50\%) \\
   & Virchow2 & 0.910 $\pm$ 0.049 & 0.915 $\pm$ 0.030 & (-1\%) & 0.958 $\pm$ 0.010 & (119\%) & 0.952 $\pm$ 0.010 & (87\%) \\
  \midrule
  \multirow{8}{*}{\rotatebox[origin=c]{90}{\parbox{2cm}{\centering Concordance\\ correlation coefficient}}}
   & H-optimus-0 & 0.475 $\pm$ 0.136 & 0.841 $\pm$ 0.067 & (227\%) & 0.976 $\pm$ 0.013 & (1914\%) & 0.961 $\pm$ 0.020 & (1258\%) \\
   & H-optimus-1 & 0.678 $\pm$ 0.102 & 0.873 $\pm$ 0.059 & (131\%) & 0.978 $\pm$ 0.007 & (1260\%) & 0.969 $\pm$ 0.018 & (945\%) \\
   & Hibou-L & 0.375 $\pm$ 0.151 & 0.844 $\pm$ 0.061 & (282\%) & 0.788 $\pm$ 0.288 & (231\%) & 0.940 $\pm$ 0.032 & (935\%) \\
   & Phikon-v2 & 0.417 $\pm$ 0.145 & 0.785 $\pm$ 0.095 & (167\%) & 0.957 $\pm$ 0.015 & (1314\%) & 0.930 $\pm$ 0.029 & (732\%) \\
   & Prov-GigaPath & 0.546 $\pm$ 0.105 & 0.800 $\pm$ 0.086 & (119\%) & 0.976 $\pm$ 0.007 & (1602\%) & 0.962 $\pm$ 0.018 & (1083\%) \\
   & UNI & 0.612 $\pm$ 0.114 & 0.859 $\pm$ 0.060 & (192\%) & 0.976 $\pm$ 0.008 & (1660\%) & 0.966 $\pm$ 0.014 & (1051\%) \\
   & UNI2-H & 0.438 $\pm$ 0.144 & 0.767 $\pm$ 0.098 & (108\%) & 0.975 $\pm$ 0.009 & (1845\%) & 0.962 $\pm$ 0.021 & (1397\%) \\
   & Virchow2 & 0.677 $\pm$ 0.127 & 0.889 $\pm$ 0.050 & (211\%) & 0.982 $\pm$ 0.013 & (1788\%) & 0.974 $\pm$ 0.011 & (1154\%) \\
  \bottomrule
\end{tabular}

\end{table}
\restoregeometry

\newgeometry{left=2cm,bottom=3cm}
\begin{table}[ht]
  \centering
  \small
  \setlength{\tabcolsep}{3pt}
  \caption{%
    \textbf{Performance comparison between different loss terms in prediction of lymph node metastasis.}
    Metric summary results are given as \emph{mean}~$\pm$~\emph{standard deviation} (\emph{relative improvement}). For all columns with robustness loss, the best weight ($\lambda$) is used, and improvement is relative to no robustness loss ($\lambda$ = 0).
  }\label{e-tab:test-lnm-improvement}
  \begin{tabular}{llrr@{\hskip 0.1cm}rr@{\hskip 0.1cm}rr@{\hskip 0.1cm}r}
  \toprule
  & \textbf{Model} & \textbf{No robustness loss} & \multicolumn{2}{c}{\textbf{Embedding loss}} & \multicolumn{2}{c}{\textbf{Score loss}} & \multicolumn{2}{c}{\textbf{Both losses}} \\
  \midrule
  \multirow{8}{*}{\rotatebox[origin=c]{90}{\parbox{2cm}{\centering Balanced accuracy}}}
   & H-optimus-0 & 0.611 $\pm$ 0.014 & 0.608 $\pm$ 0.014 & (0\%) & 0.647 $\pm$ 0.014 & (10\%) & 0.620 $\pm$ 0.019 & (2\%) \\
   & H-optimus-1 & 0.631 $\pm$ 0.009 & 0.638 $\pm$ 0.011 & (5\%) & 0.628 $\pm$ 0.018 & (2\%) & 0.612 $\pm$ 0.026 & (-5\%) \\
   & Hibou-L & 0.661 $\pm$ 0.013 & 0.655 $\pm$ 0.013 & (2\%) & 0.662 $\pm$ 0.013 & (6\%) & 0.688 $\pm$ 0.016 & (8\%) \\
   & Phikon-v2 & 0.618 $\pm$ 0.011 & 0.617 $\pm$ 0.014 & (6\%) & 0.625 $\pm$ 0.015 & (8\%) & 0.620 $\pm$ 0.008 & (0\%) \\
   & Prov-GigaPath & 0.635 $\pm$ 0.011 & 0.620 $\pm$ 0.009 & (0\%) & 0.654 $\pm$ 0.011 & (10\%) & 0.651 $\pm$ 0.015 & (4\%) \\
   & UNI & 0.636 $\pm$ 0.011 & 0.640 $\pm$ 0.014 & (8\%) & 0.653 $\pm$ 0.014 & (14\%) & 0.605 $\pm$ 0.013 & (-8\%) \\
   & UNI2-H & 0.622 $\pm$ 0.011 & 0.626 $\pm$ 0.009 & (5\%) & 0.653 $\pm$ 0.011 & (13\%) & 0.637 $\pm$ 0.016 & (4\%) \\
   & Virchow2 & 0.602 $\pm$ 0.016 & 0.633 $\pm$ 0.010 & (8\%) & 0.573 $\pm$ 0.015 & (-7\%) & 0.564 $\pm$ 0.015 & (-9\%) \\
  \midrule
  \multirow{8}{*}{\rotatebox[origin=c]{90}{\parbox{2cm}{\centering AUC}}}
   & H-optimus-0 & 0.655 $\pm$ 0.012 & 0.648 $\pm$ 0.012 & (1\%) & 0.727 $\pm$ 0.005 & (29\%) & 0.732 $\pm$ 0.004 & (28\%) \\
   & H-optimus-1 & 0.662 $\pm$ 0.007 & 0.669 $\pm$ 0.011 & (4\%) & 0.729 $\pm$ 0.008 & (27\%) & 0.737 $\pm$ 0.004 & (28\%) \\
   & Hibou-L & 0.722 $\pm$ 0.016 & 0.714 $\pm$ 0.008 & (3\%) & 0.749 $\pm$ 0.005 & (20\%) & 0.758 $\pm$ 0.007 & (14\%) \\
   & Phikon-v2 & 0.673 $\pm$ 0.011 & 0.667 $\pm$ 0.013 & (8\%) & 0.706 $\pm$ 0.004 & (23\%) & 0.707 $\pm$ 0.004 & (11\%) \\
   & Prov-GigaPath & 0.685 $\pm$ 0.009 & 0.670 $\pm$ 0.008 & (3\%) & 0.728 $\pm$ 0.004 & (24\%) & 0.726 $\pm$ 0.003 & (15\%) \\
   & UNI & 0.698 $\pm$ 0.010 & 0.688 $\pm$ 0.012 & (8\%) & 0.732 $\pm$ 0.008 & (28\%) & 0.726 $\pm$ 0.004 & (10\%) \\
   & UNI2-H & 0.663 $\pm$ 0.013 & 0.667 $\pm$ 0.007 & (8\%) & 0.735 $\pm$ 0.004 & (35\%) & 0.742 $\pm$ 0.004 & (30\%) \\
   & Virchow2 & 0.630 $\pm$ 0.016 & 0.668 $\pm$ 0.007 & (11\%) & 0.728 $\pm$ 0.003 & (34\%) & 0.730 $\pm$ 0.007 & (36\%) \\
  \midrule
  \multirow{8}{*}{\rotatebox[origin=c]{90}{\parbox{2cm}{\centering Combined score}}}
   & H-optimus-0 & 0.550 $\pm$ 0.012 & 0.556 $\pm$ 0.011 & (4\%) & 0.672 $\pm$ 0.006 & (41\%) & 0.677 $\pm$ 0.005 & (39\%) \\
   & H-optimus-1 & 0.559 $\pm$ 0.011 & 0.589 $\pm$ 0.015 & (10\%) & 0.684 $\pm$ 0.006 & (43\%) & 0.692 $\pm$ 0.005 & (43\%) \\
   & Hibou-L & 0.591 $\pm$ 0.021 & 0.650 $\pm$ 0.008 & (27\%) & 0.680 $\pm$ 0.006 & (44\%) & 0.691 $\pm$ 0.007 & (32\%) \\
   & Phikon-v2 & 0.526 $\pm$ 0.019 & 0.558 $\pm$ 0.019 & (20\%) & 0.636 $\pm$ 0.005 & (45\%) & 0.644 $\pm$ 0.004 & (33\%) \\
   & Prov-GigaPath & 0.579 $\pm$ 0.010 & 0.578 $\pm$ 0.009 & (6\%) & 0.663 $\pm$ 0.009 & (32\%) & 0.667 $\pm$ 0.005 & (26\%) \\
   & UNI & 0.523 $\pm$ 0.019 & 0.530 $\pm$ 0.015 & (12\%) & 0.632 $\pm$ 0.012 & (44\%) & 0.650 $\pm$ 0.006 & (36\%) \\
   & UNI2-H & 0.554 $\pm$ 0.019 & 0.579 $\pm$ 0.009 & (12\%) & 0.690 $\pm$ 0.004 & (52\%) & 0.694 $\pm$ 0.005 & (45\%) \\
   & Virchow2 & 0.492 $\pm$ 0.026 & 0.561 $\pm$ 0.012 & (15\%) & 0.672 $\pm$ 0.004 & (55\%) & 0.677 $\pm$ 0.007 & (57\%) \\
  \midrule
  \multirow{8}{*}{\rotatebox[origin=c]{90}{\parbox{2cm}{\centering Inconsistency}}}
   & H-optimus-0 & 0.167 $\pm$ 0.008 & 0.146 $\pm$ 0.009 & (18\%) & 0.087 $\pm$ 0.005 & (100\%) & 0.088 $\pm$ 0.004 & (90\%) \\
   & H-optimus-1 & 0.164 $\pm$ 0.013 & 0.128 $\pm$ 0.012 & (34\%) & 0.071 $\pm$ 0.007 & (136\%) & 0.071 $\pm$ 0.004 & (131\%) \\
   & Hibou-L & 0.210 $\pm$ 0.016 & 0.102 $\pm$ 0.006 & (137\%) & 0.110 $\pm$ 0.006 & (130\%) & 0.105 $\pm$ 0.004 & (98\%) \\
   & Phikon-v2 & 0.235 $\pm$ 0.024 & 0.175 $\pm$ 0.016 & (53\%) & 0.112 $\pm$ 0.005 & (135\%) & 0.100 $\pm$ 0.005 & (134\%) \\
   & Prov-GigaPath & 0.169 $\pm$ 0.009 & 0.146 $\pm$ 0.009 & (16\%) & 0.104 $\pm$ 0.012 & (63\%) & 0.094 $\pm$ 0.006 & (79\%) \\
   & UNI & 0.279 $\pm$ 0.021 & 0.252 $\pm$ 0.018 & (21\%) & 0.159 $\pm$ 0.012 & (88\%) & 0.122 $\pm$ 0.006 & (129\%) \\
   & UNI2-H & 0.173 $\pm$ 0.016 & 0.141 $\pm$ 0.012 & (30\%) & 0.072 $\pm$ 0.003 & (149\%) & 0.076 $\pm$ 0.004 & (127\%) \\
   & Virchow2 & 0.221 $\pm$ 0.024 & 0.171 $\pm$ 0.013 & (29\%) & 0.089 $\pm$ 0.005 & (157\%) & 0.084 $\pm$ 0.006 & (163\%) \\
  \midrule
  \multirow{8}{*}{\rotatebox[origin=c]{90}{\parbox{2cm}{\centering Classification\\ agreement}}}
   & H-optimus-0 & 0.849 $\pm$ 0.017 & 0.866 $\pm$ 0.021 & (15\%) & 0.943 $\pm$ 0.014 & (178\%) & 0.949 $\pm$ 0.010 & (197\%) \\
   & H-optimus-1 & 0.848 $\pm$ 0.019 & 0.889 $\pm$ 0.020 & (42\%) & 0.952 $\pm$ 0.014 & (213\%) & 0.961 $\pm$ 0.013 & (293\%) \\
   & Hibou-L & 0.830 $\pm$ 0.020 & 0.915 $\pm$ 0.013 & (128\%) & 0.930 $\pm$ 0.009 & (200\%) & 0.921 $\pm$ 0.012 & (114\%) \\
   & Phikon-v2 & 0.800 $\pm$ 0.022 & 0.847 $\pm$ 0.021 & (53\%) & 0.922 $\pm$ 0.014 & (198\%) & 0.926 $\pm$ 0.012 & (171\%) \\
   & Prov-GigaPath & 0.857 $\pm$ 0.018 & 0.876 $\pm$ 0.014 & (21\%) & 0.923 $\pm$ 0.011 & (87\%) & 0.927 $\pm$ 0.010 & (96\%) \\
   & UNI & 0.752 $\pm$ 0.023 & 0.769 $\pm$ 0.020 & (19\%) & 0.880 $\pm$ 0.020 & (123\%) & 0.919 $\pm$ 0.012 & (204\%) \\
   & UNI2-H & 0.842 $\pm$ 0.021 & 0.875 $\pm$ 0.021 & (31\%) & 0.951 $\pm$ 0.009 & (228\%) & 0.960 $\pm$ 0.008 & (297\%) \\
   & Virchow2 & 0.799 $\pm$ 0.032 & 0.857 $\pm$ 0.018 & (40\%) & 0.970 $\pm$ 0.014 & (588\%) & 0.971 $\pm$ 0.009 & (598\%) \\
  \midrule
  \multirow{8}{*}{\rotatebox[origin=c]{90}{\parbox{2cm}{\centering Concordance\\ correlation coefficient}}}
   & H-optimus-0 & 0.835 $\pm$ 0.016 & 0.883 $\pm$ 0.011 & (40\%) & 0.962 $\pm$ 0.004 & (351\%) & 0.965 $\pm$ 0.003 & (365\%) \\
   & H-optimus-1 & 0.835 $\pm$ 0.020 & 0.912 $\pm$ 0.014 & (86\%) & 0.973 $\pm$ 0.004 & (492\%) & 0.975 $\pm$ 0.002 & (548\%) \\
   & Hibou-L & 0.791 $\pm$ 0.031 & 0.948 $\pm$ 0.007 & (359\%) & 0.948 $\pm$ 0.006 & (399\%) & 0.947 $\pm$ 0.004 & (295\%) \\
   & Phikon-v2 & 0.718 $\pm$ 0.044 & 0.852 $\pm$ 0.025 & (125\%) & 0.938 $\pm$ 0.004 & (422\%) & 0.949 $\pm$ 0.004 & (456\%) \\
   & Prov-GigaPath & 0.824 $\pm$ 0.014 & 0.880 $\pm$ 0.011 & (38\%) & 0.950 $\pm$ 0.008 & (221\%) & 0.959 $\pm$ 0.004 & (332\%) \\
   & UNI & 0.658 $\pm$ 0.036 & 0.744 $\pm$ 0.026 & (39\%) & 0.905 $\pm$ 0.013 & (269\%) & 0.945 $\pm$ 0.004 & (519\%) \\
   & UNI2-H & 0.828 $\pm$ 0.025 & 0.894 $\pm$ 0.017 & (64\%) & 0.974 $\pm$ 0.001 & (557\%) & 0.976 $\pm$ 0.002 & (615\%) \\
   & Virchow2 & 0.763 $\pm$ 0.039 & 0.873 $\pm$ 0.017 & (78\%) & 0.969 $\pm$ 0.002 & (658\%) & 0.972 $\pm$ 0.003 & (752\%) \\
  \bottomrule
\end{tabular}

\end{table}
\restoregeometry

\newgeometry{top=0.5cm,bottom=0cm}
\begin{figure}[ht]
  \centering
  \includegraphics[width=.9\textwidth]{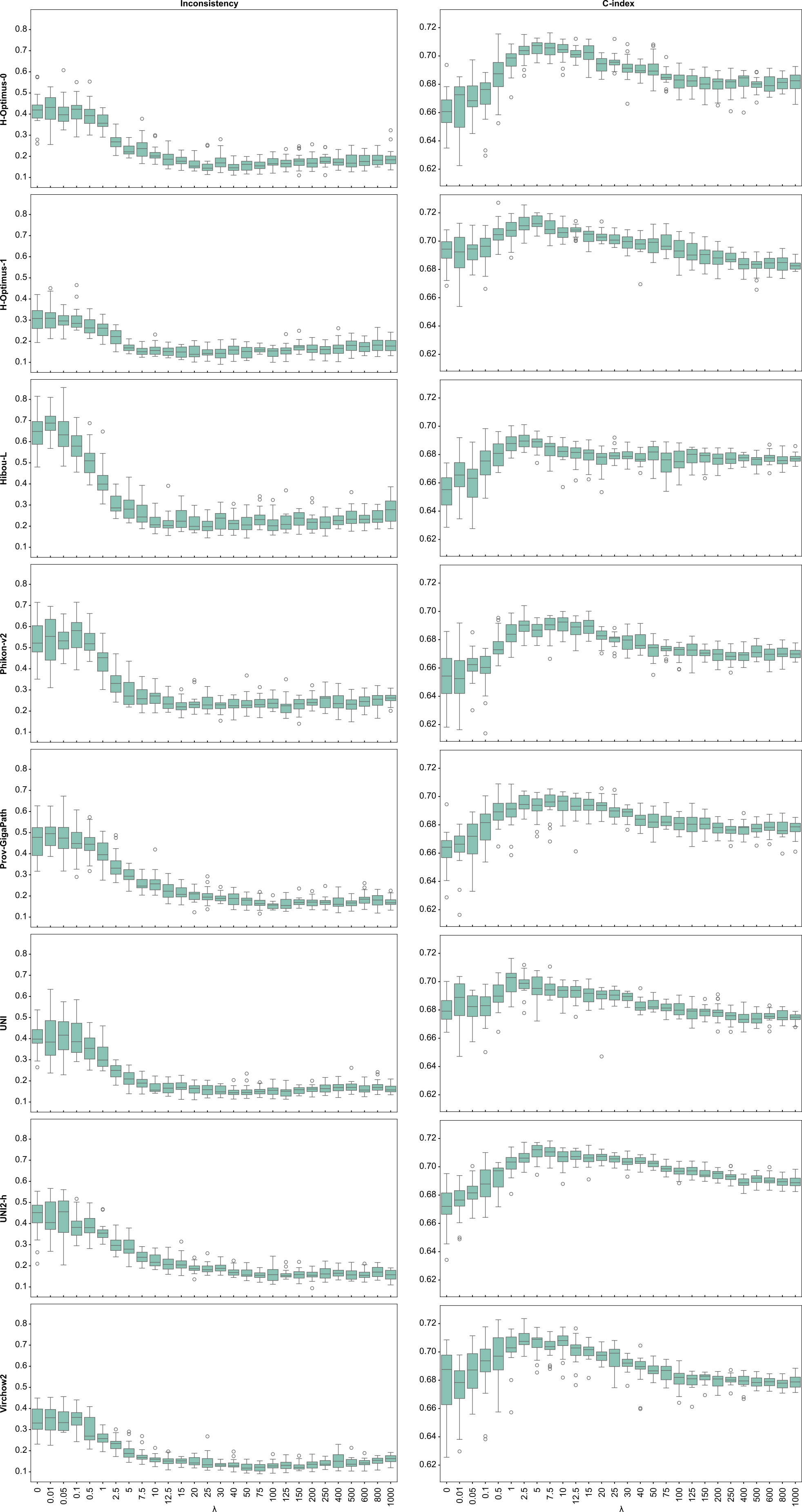}
  \caption{%
    \textbf{Performance vs robustness weight for outcome prediction.}
    Effect of varying $\lambda$ in \cref{eq:total-loss} on inconsistency and c-index. Statistics are gathered from 20 models per value of $\lambda$ per foundation model.
  }\label{e-fig:metric-vs-weight_outcome}
\end{figure}

\begin{figure}[ht]
  \centering
  \includegraphics[width=.9\textwidth]{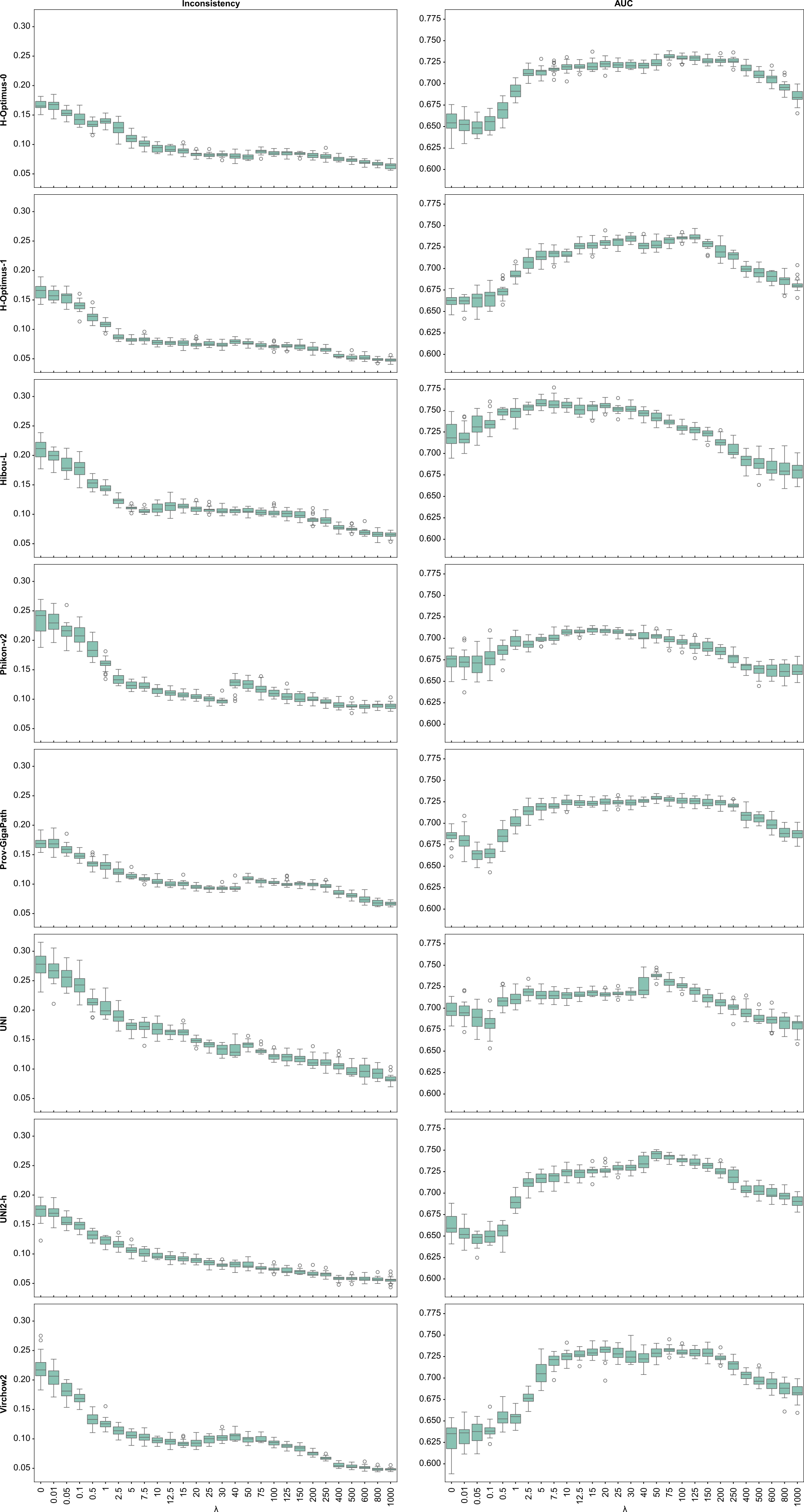}
  \caption{%
    \textbf{Performance vs robustness weight for LNM prediction.}
    Effect of varying $\lambda$ in \cref{eq:total-loss} on inconsistency and AUC. Statistics are gathered from 20 models per value of $\lambda$ per foundation model.
  }\label{e-fig:metric-vs-weight_lnm}
\end{figure}
\restoregeometry

\newgeometry{bottom=3cm}
\begin{figure}[ht]
  \centering
  \includegraphics[width=0.9\textwidth]{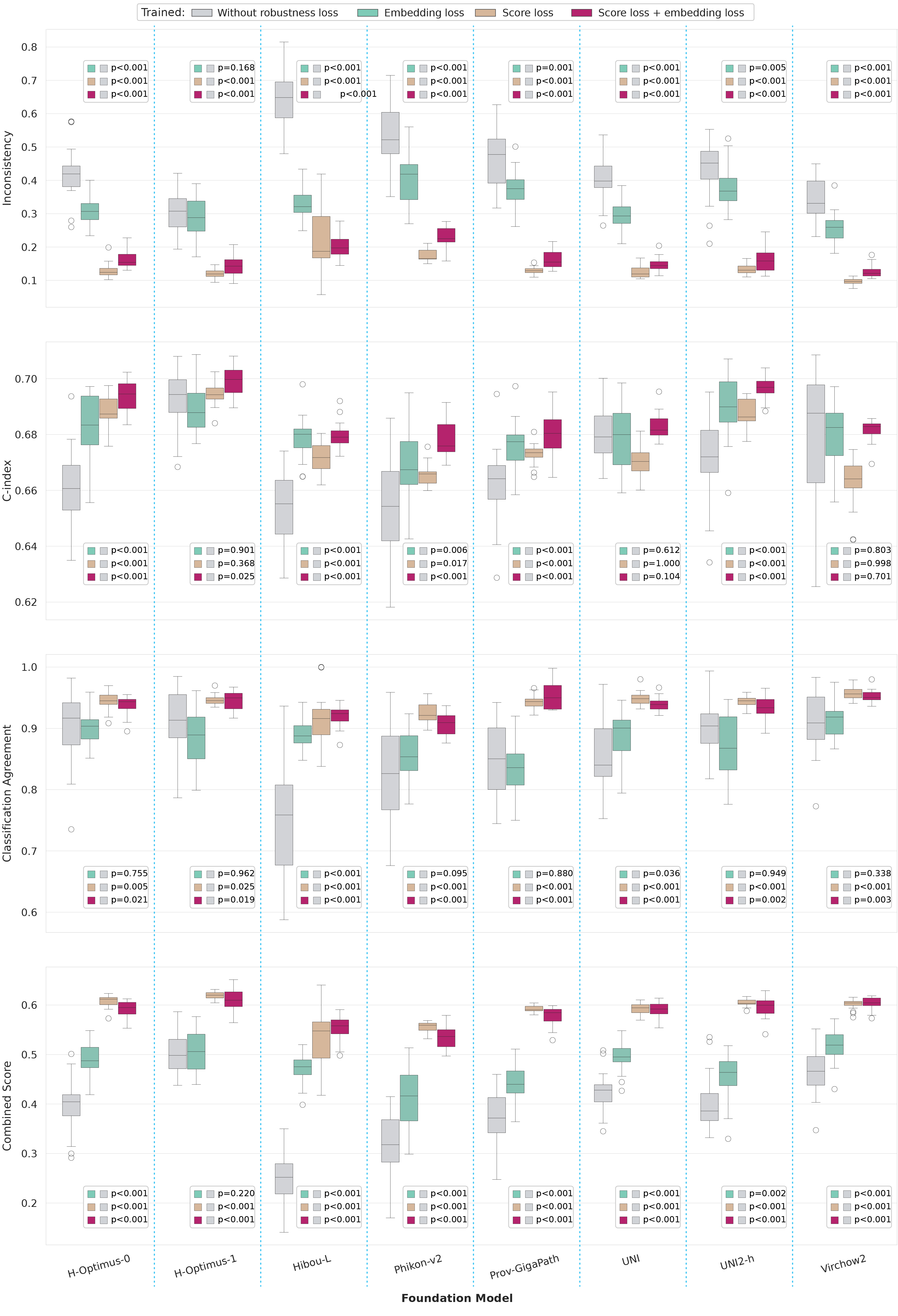}
  \caption{%
    \textbf{Effect of different robustness loss terms on robustness and classification performance in survival prediction.}
    Box plots show the distribution of metric values for each foundation model when downstream models are trained without robustness loss, with the embedding loss alone, with the score loss alone, or with both loss terms combined. P values indicate comparisons of each loss setting with conventional training without robustness loss.
  }\label{e-fig:outcome_model_comparison_all_3}
\end{figure}

\begin{figure}[ht]
  \centering
  \includegraphics[width=0.9\textwidth]{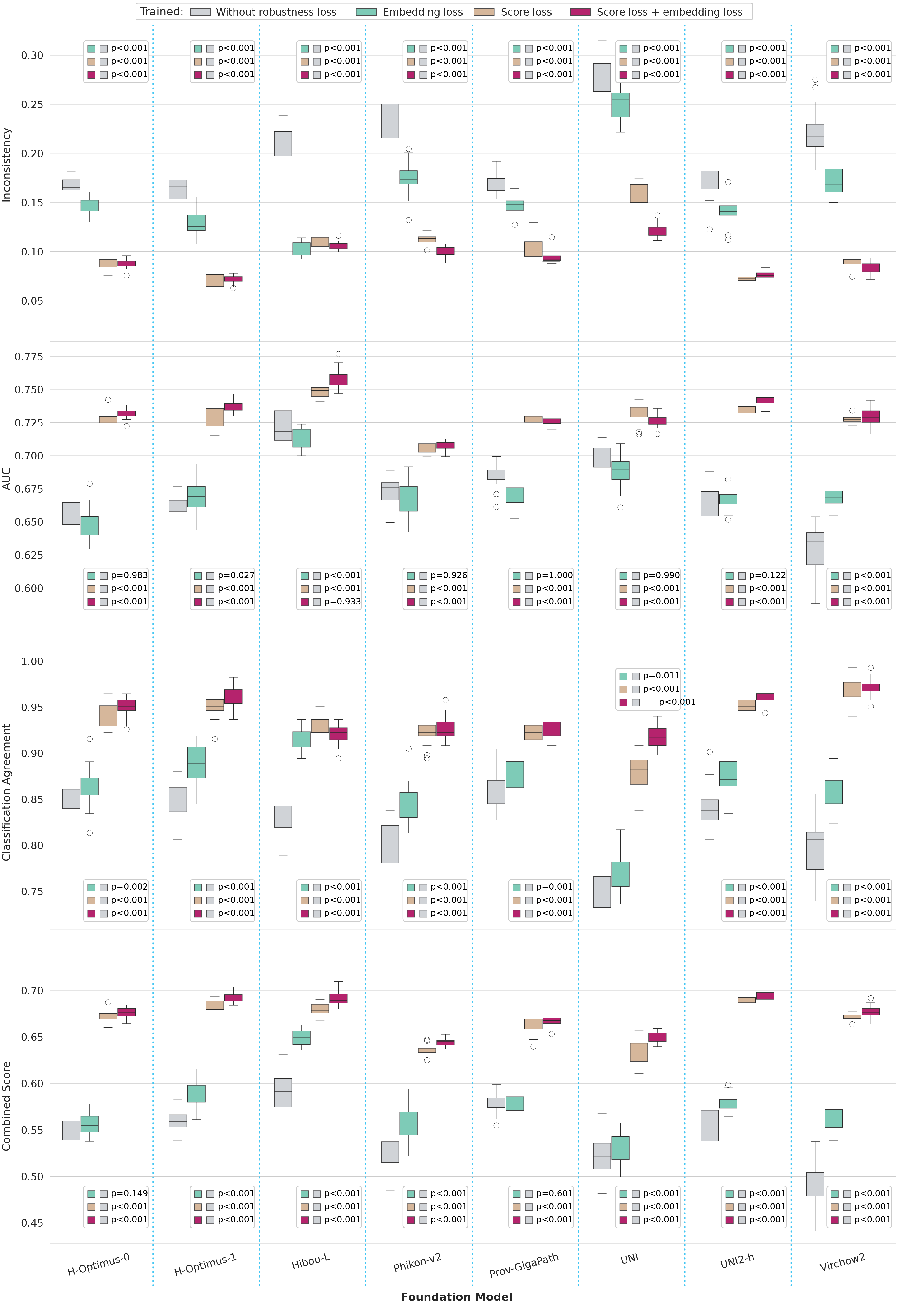}
  \caption{%
    \textbf{Effect of different robustness loss terms on robustness and classification performance in prediction of lymph node metastasis.}
    Box plots show the distribution of metric values for each foundation model when downstream models are trained without robustness loss, with the embedding loss alone, with the score loss alone, or with both loss terms combined. P values indicate comparisons of each loss setting with conventional training without robustness loss.
  }\label{e-fig:lnm_model_comparison_all_3}
\end{figure}
\restoregeometry

\begin{figure}[ht]
  \centering
  \includegraphics[width=\textwidth]{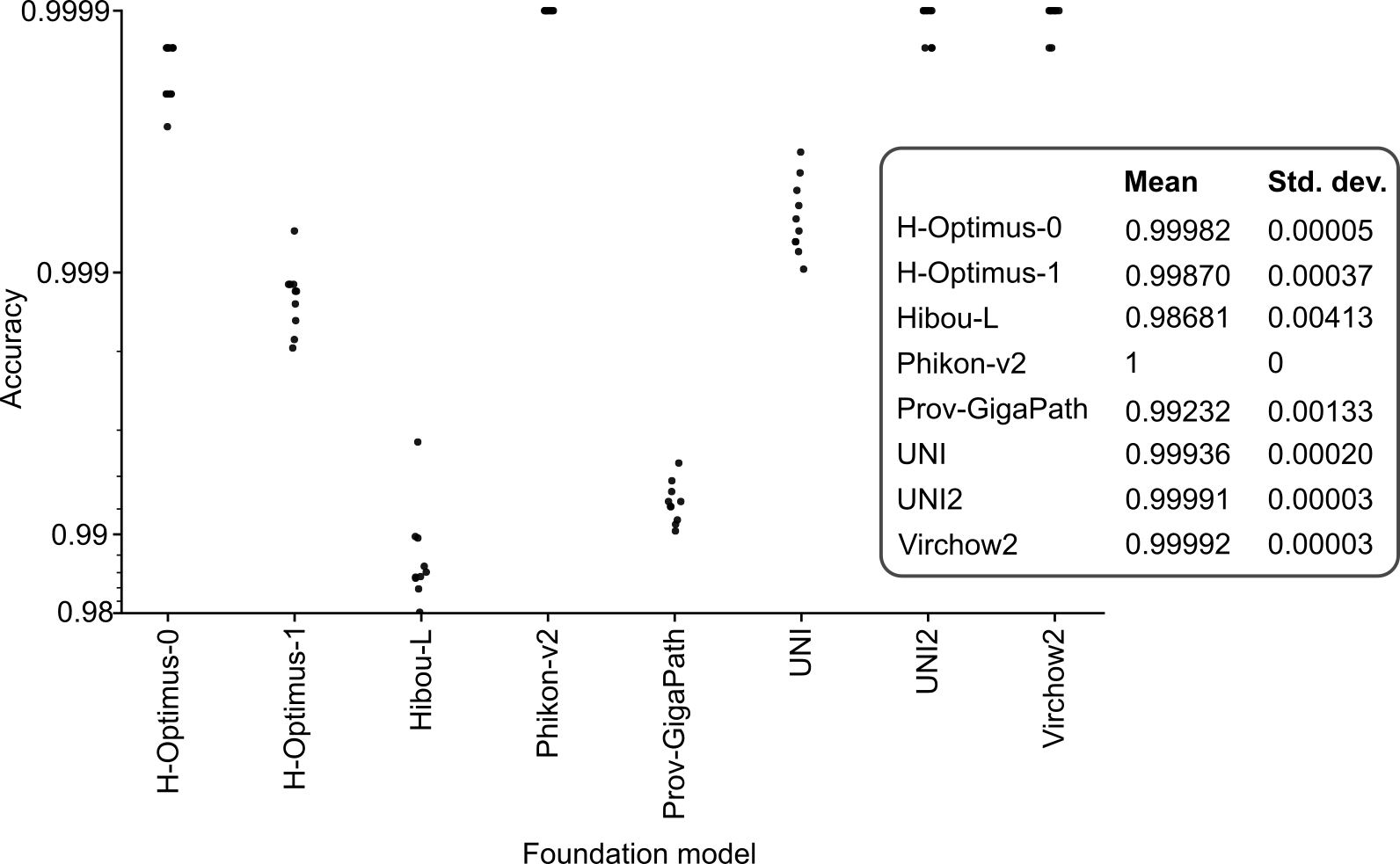}
  \caption{%
    \textbf{Scanner prediction with linear probing.}
    A single-layer linear classifier was trained to predict which scanner was used to form the input WSI, using tile feature vectors extracted with foundation models.
    Input is a WSI feature vector that is the average of its tile feature vectors, and the task was to predict whether a WSI had been imaged with one of the following five scanners: Aperio AT2, Aperio GT 450 DX, NanoZoomer XR, KF-PRO-400, Pannoramic 1000.
    We trained nine classifiers for each of the eight foundation models on WSIs from QUASAR 2 and show the average accuracy over all scanners on WSIs from TransSCOT in this figure.
    We used PyTorch's implementation of the LBFGS optimisation algorithm with a learning rate of 1.0, 1000 as the maximum number of iterations, a history size of 50, with strong wolfe as the solver, and CrossEntropyLoss as the loss function.
    Prior to feeding each feature vector into the linear classifier, we normalise it based on mean and standard deviation calculated on the QUASAR 2 dataset.
  }\label{e-fig:linear-probing}
\end{figure}

\clearpage
\appendix


\section*{Supplementary material}

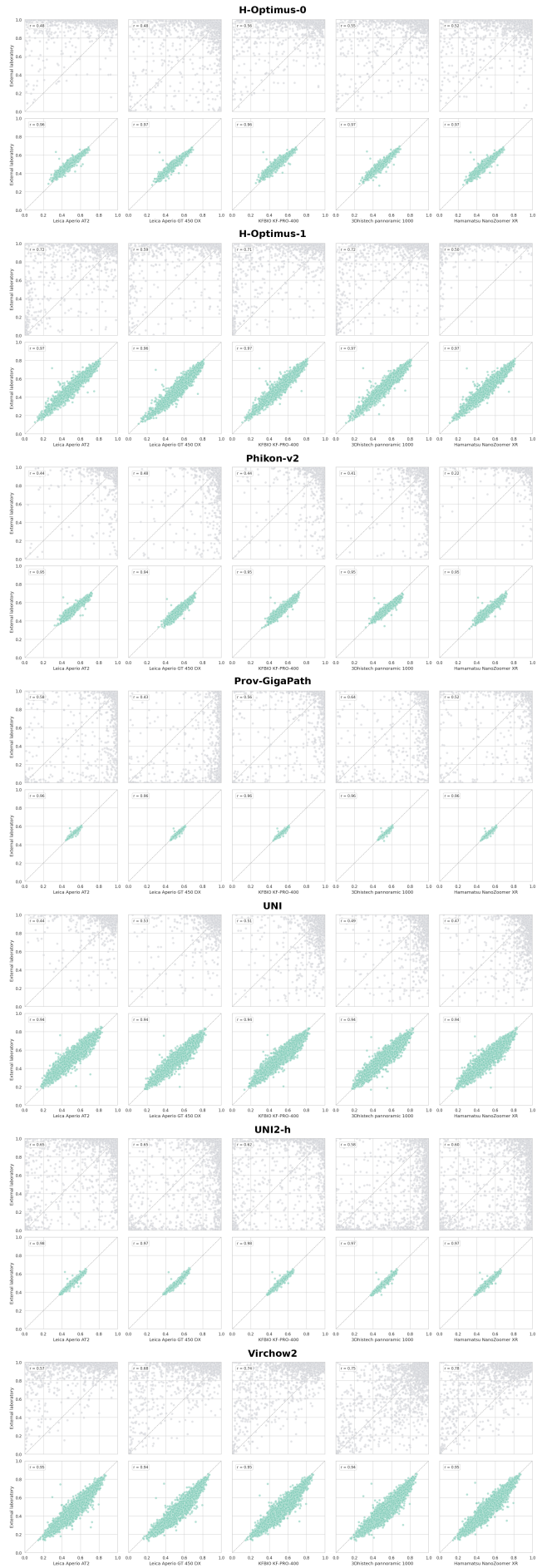

**Extended Data Fig. 1 | WSI prediction score comparison between tissue sections**
Like scatterplots in Fig. 2 for the remaining seven foundation models.

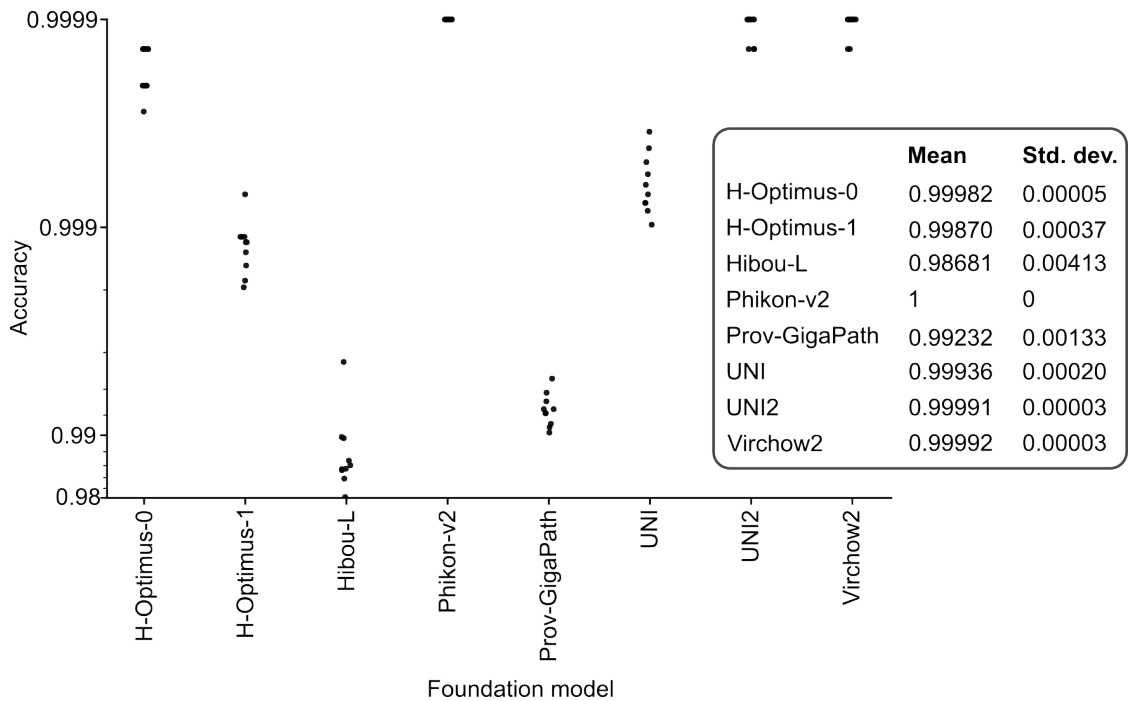

**Extended Data Fig. 2 | Scanner prediction with linear probing**
A linear classifier was trained on top of raw features from eight different foundation models for computational pathology. The classifier was trained on WSIs from QUASAR 2 and applied on WSIs from TransSCOT. The task was to predict whether a scan had been imaged with one of the following five scanners: Aperio AT2, Aperio GT 450 DX, NanoZoomer XR, KF-PRO-400, P1000.

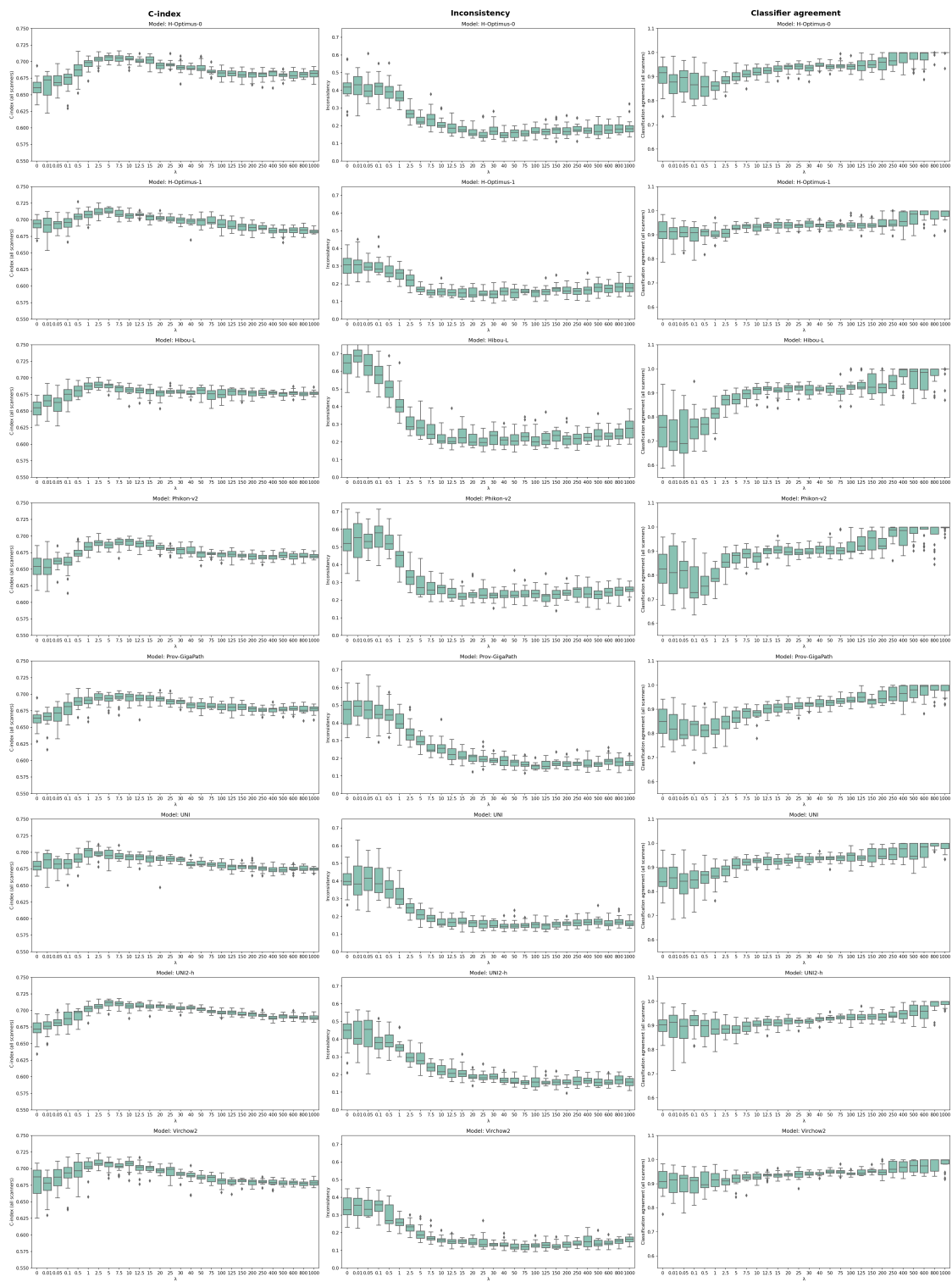

**Extended Data Fig. 3 | Performance with different loss weights on TransSCOT**

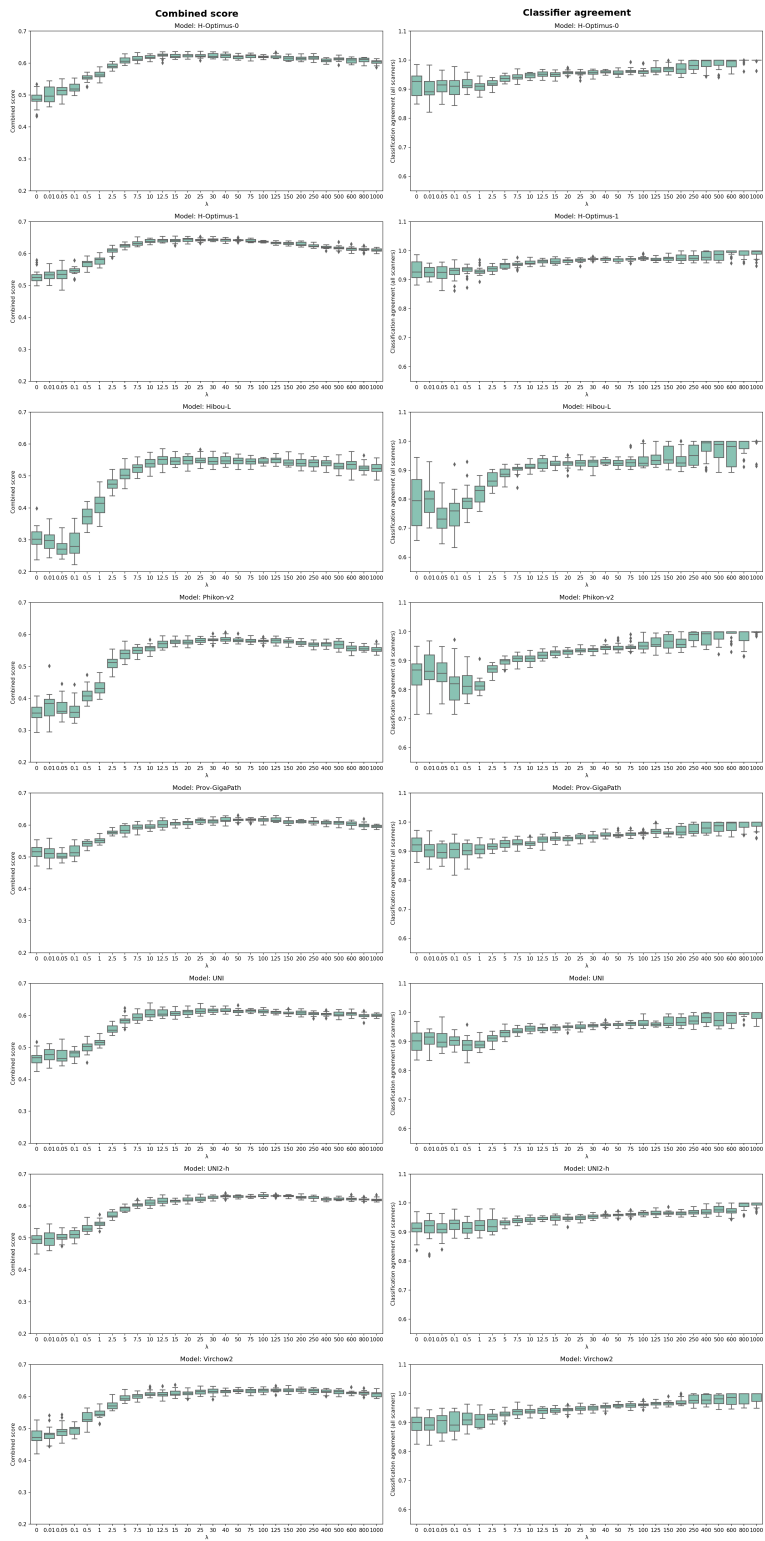

**Extended Data Fig. 4 | Performance with different loss weights on QUASAR 2**

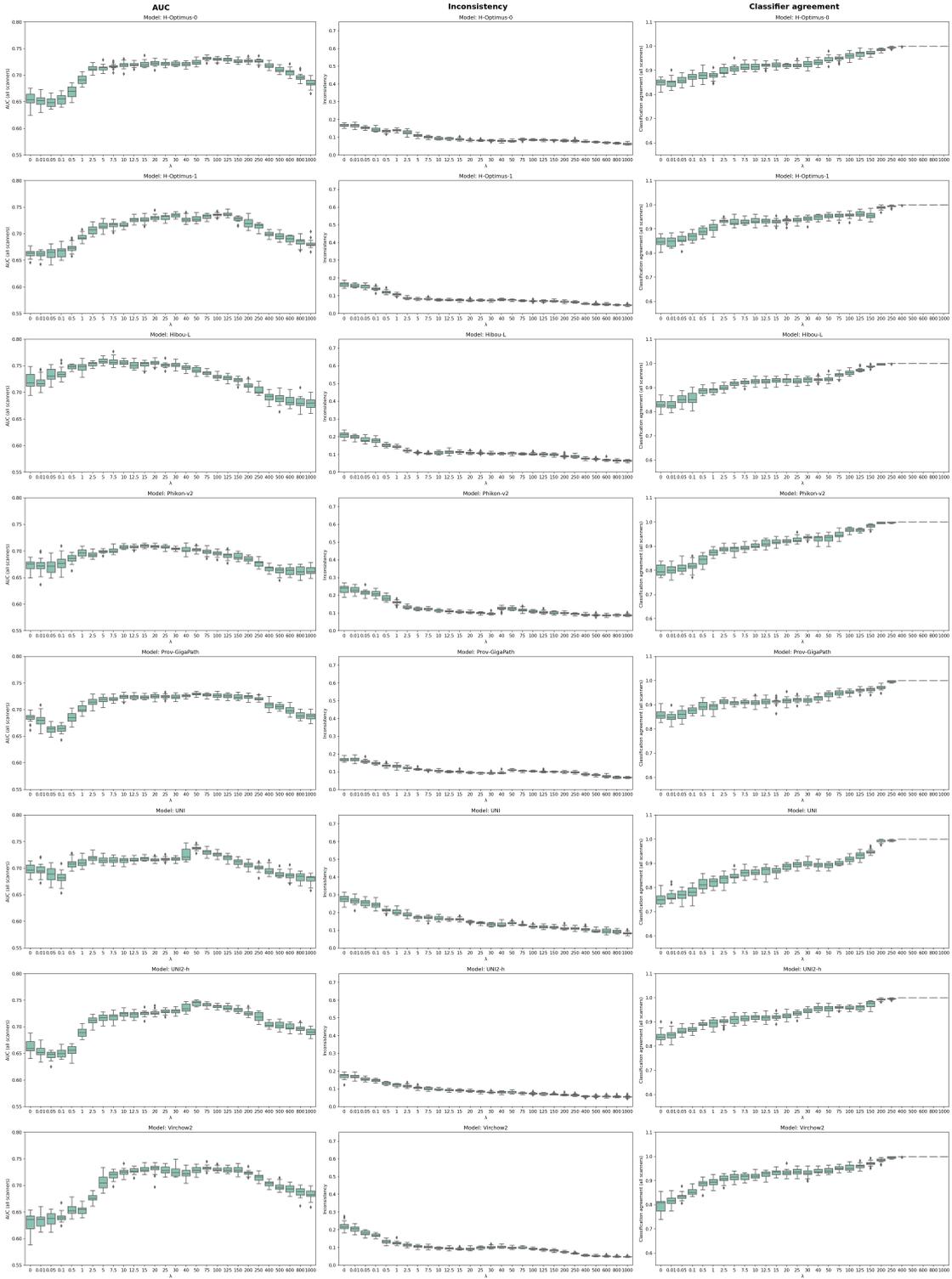

**Extended Data Fig. 5 | Performance with different loss weights on Dutch T1**

# Enabling clinical use of foundation models in histopathology

**Supplementary Information**

# 1 Materials

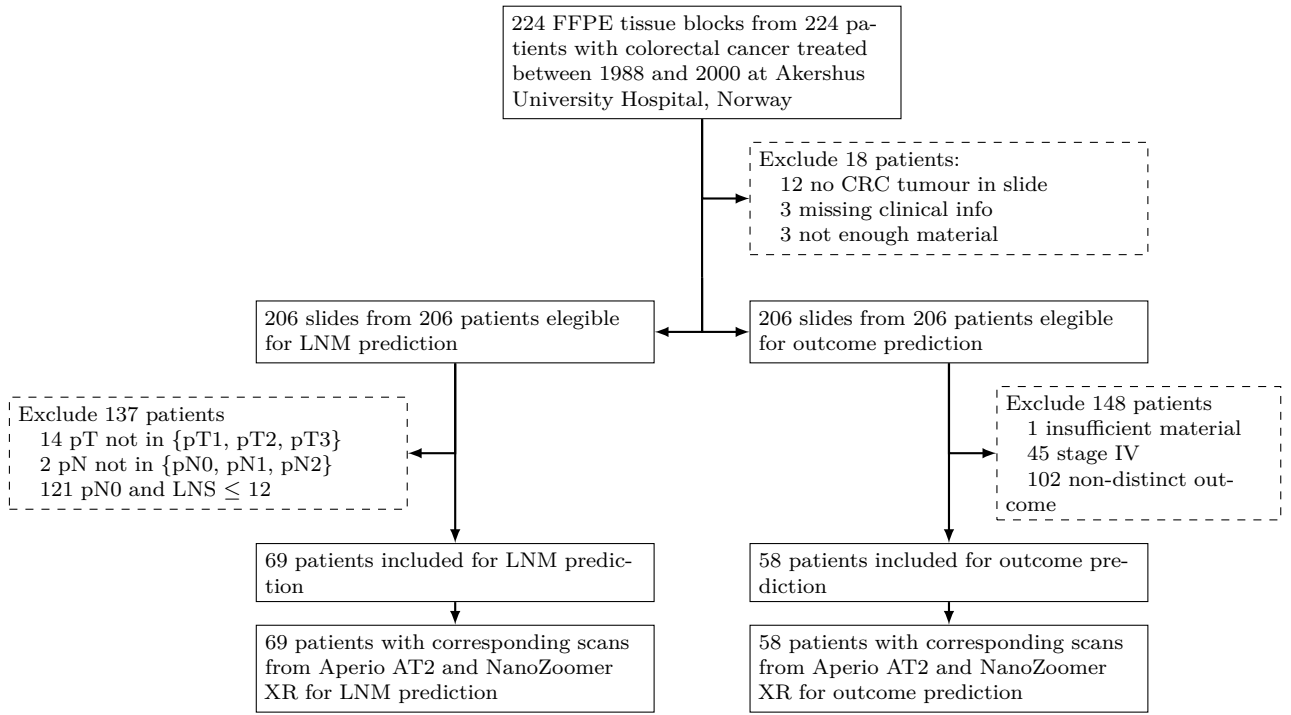

**Supplementary Fig. 1** | Flow from received blocks to corresponding scans for the Ahus cohort

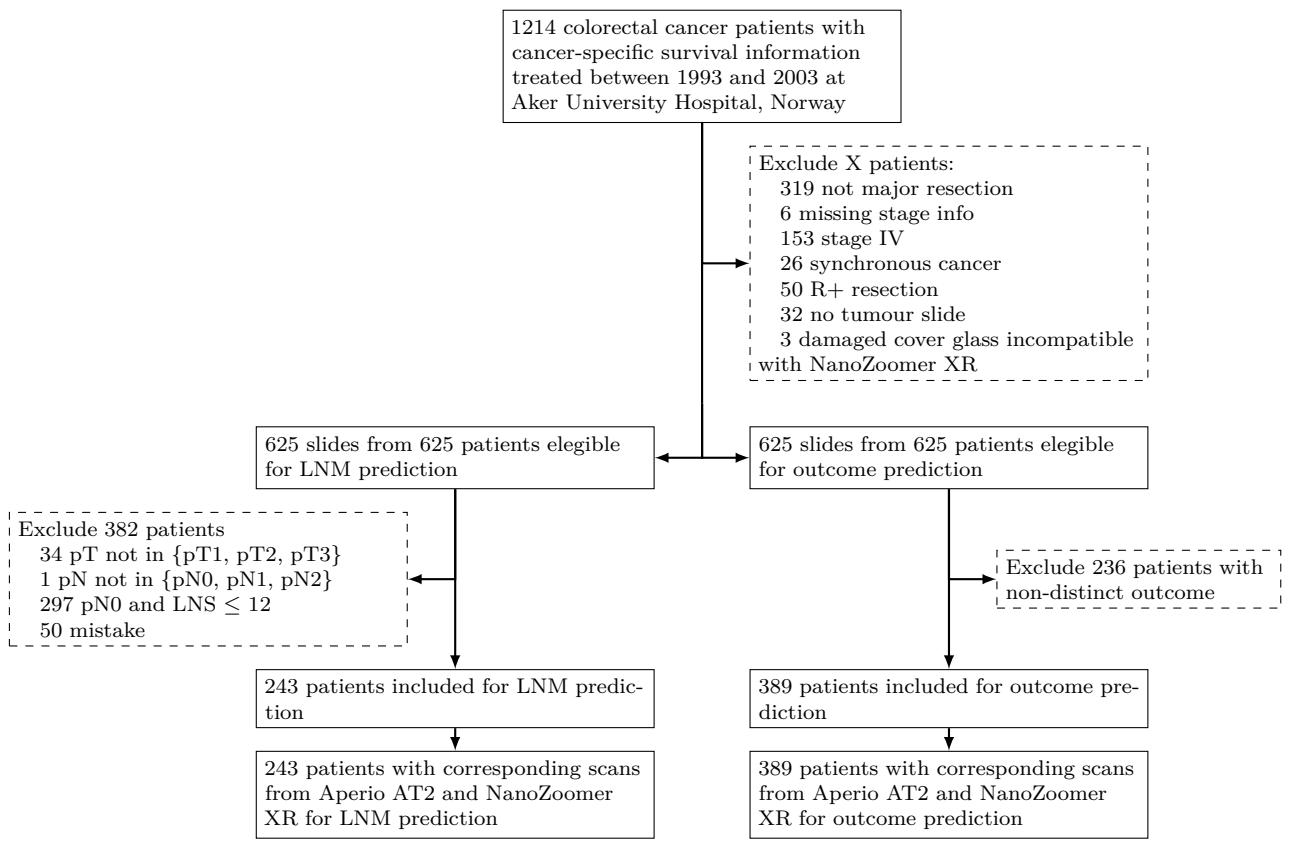

**Supplementary Fig. 2** | Flow from received blocks to corresponding scans for the Aker cohort

53

2

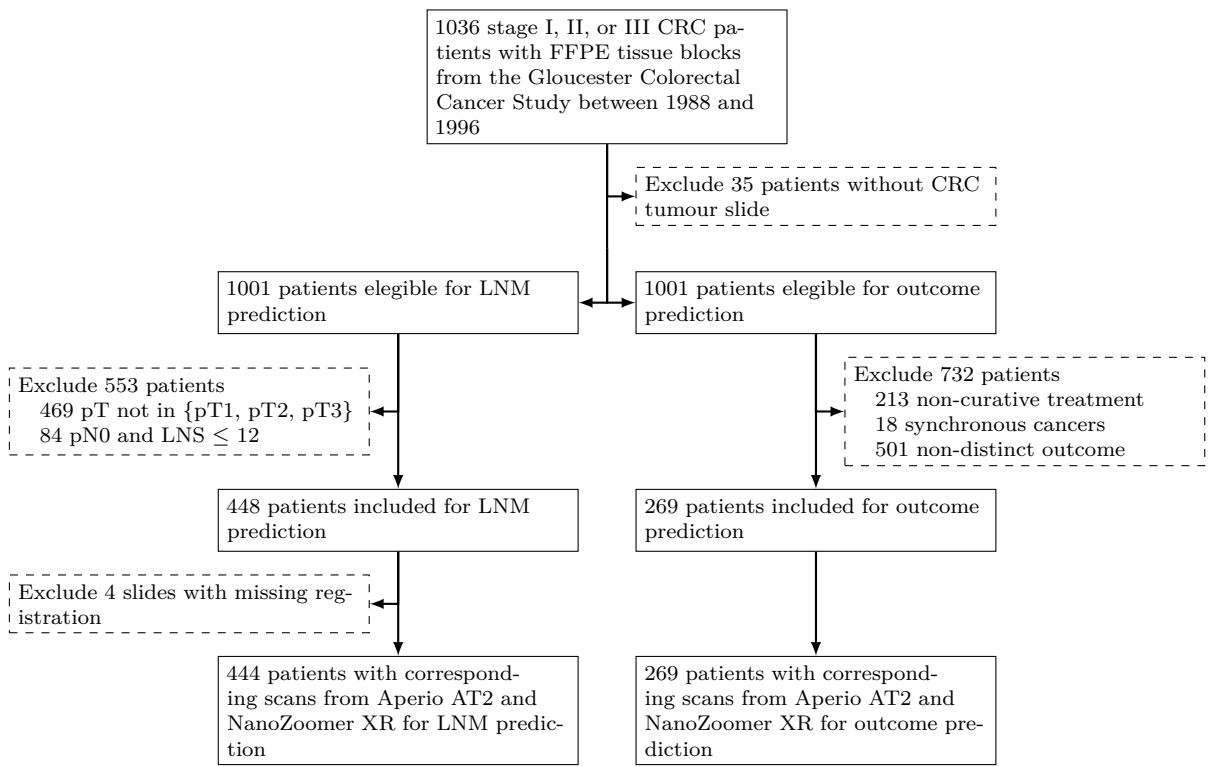

**Supplementary Fig. 3** | Flow from original study recruitment to corresponding scans for the Gloucester cohort

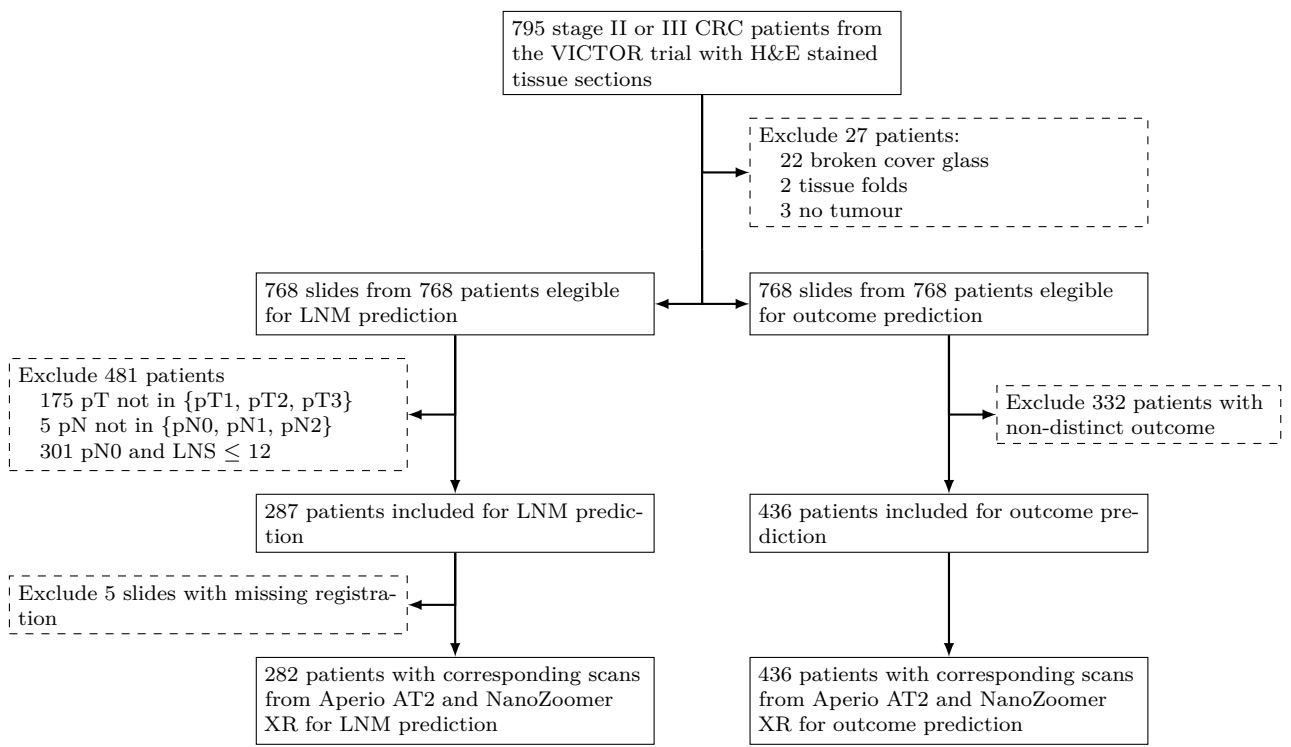

**Supplementary Fig. 4** | Flow from received slides to annotated scans for the Victor cohort

3

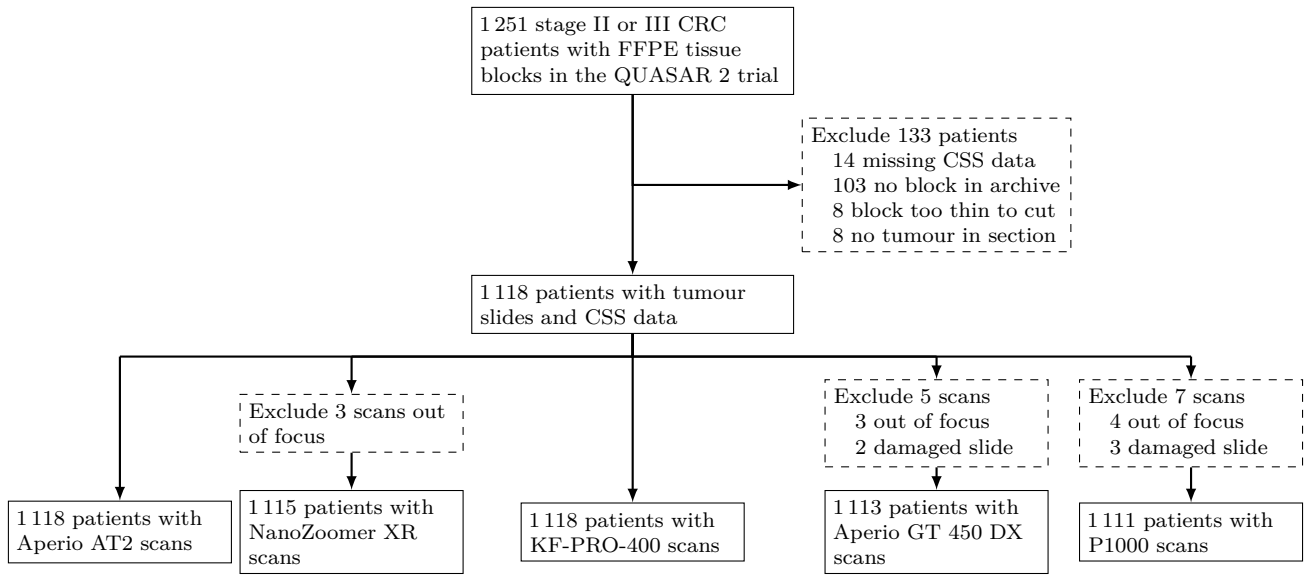

**Supplementary Fig. 5** | Flow from received blocks to scans for the QUASAR 2 cohort

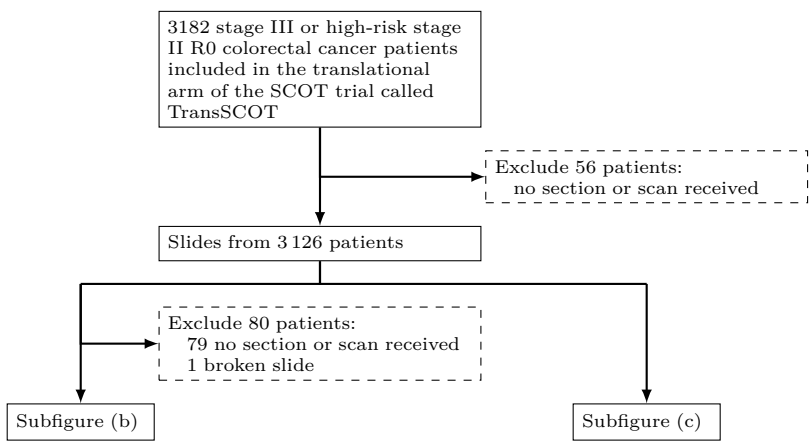

**(a)** Flow from included patients to included scans for the TransSCOT cohort

55

4

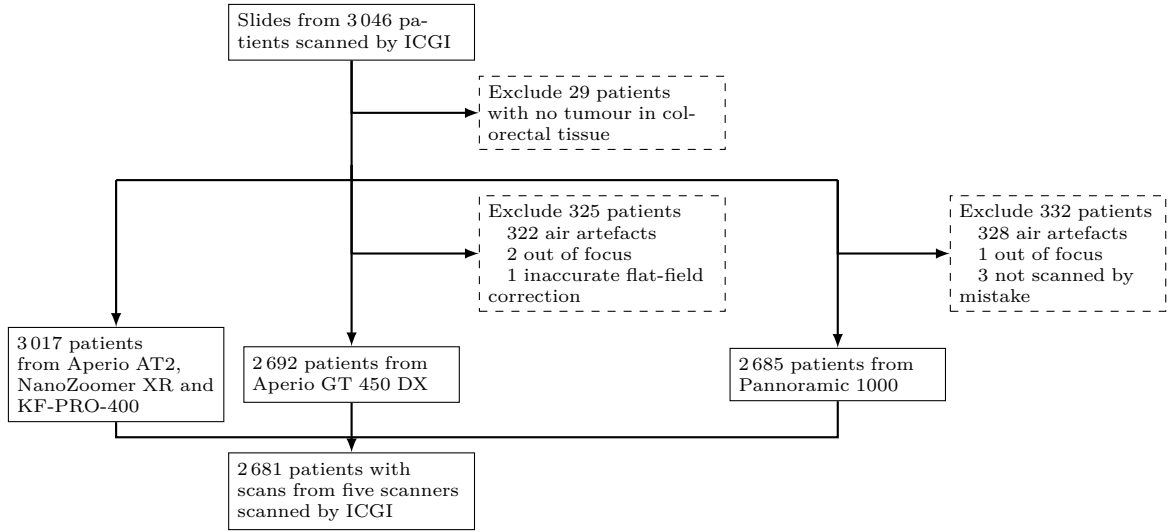

**(b)** Flow from included scans to scans included from sections received by ICGI

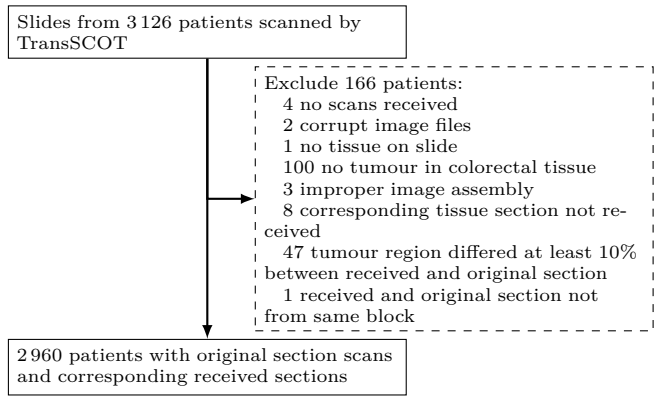

**(c)** Flow from included scans to scans included from original sections

**Supplementary Fig. 6** | Flow from included patients to included scans for the TransSCOT cohort

56

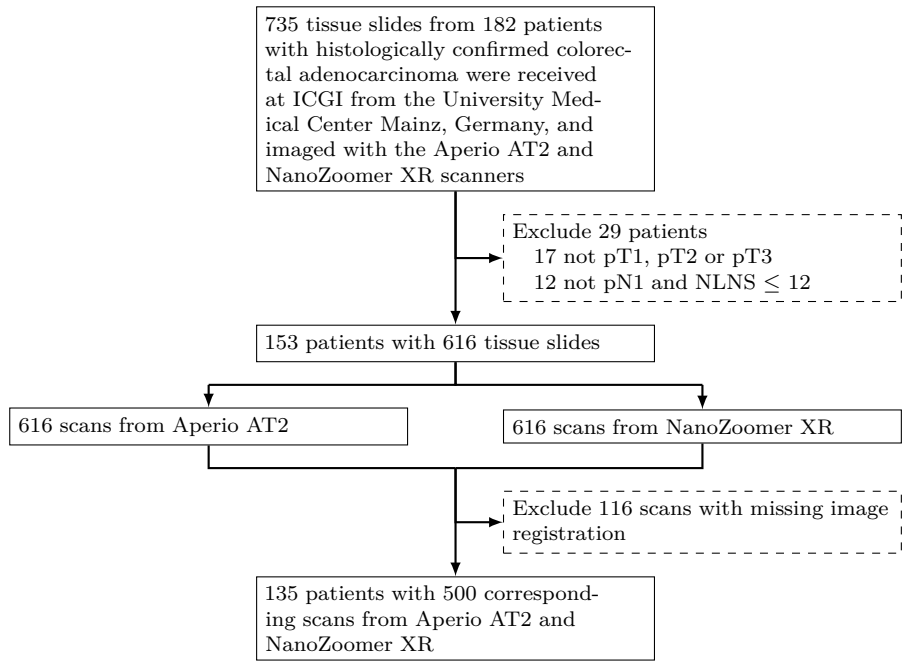

**Supplementary Fig. 7** | Flow from received slides to corresponding scans for the Mainz cohort

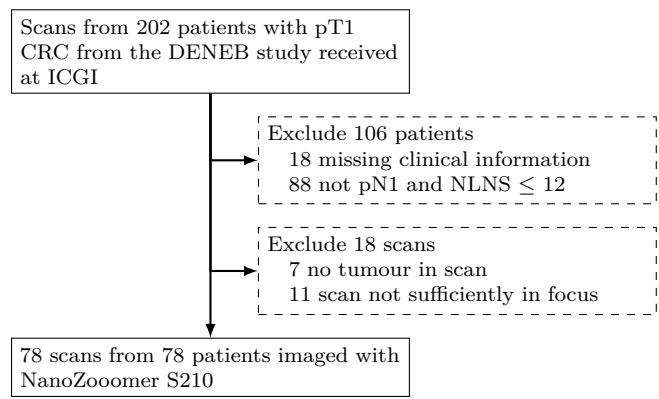

**Supplementary Fig. 8** | Flow from received to included scans in the DENEB cohort

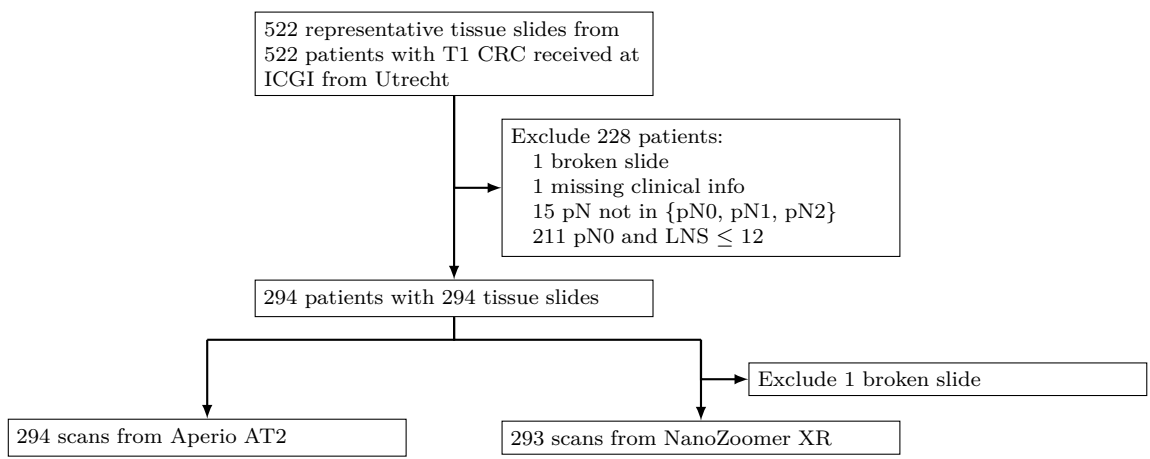

**Supplementary Fig. 9** | Flow from received slides to annotated scans for the Dutch T1 cohort

**Supplementary Table 1 | Patient and scan count outcome prediction**

| Cohort | Patients | Scans | | | | | | |
|---|---|---|---|---|---|---|---|---|
| | | Aperio AT2 | NanoZoomer XR | KF-PRO-400 | Aperio GT 450 DX | P1000 | Original | Sum |
| Ahus | 58 | 58 | 58 | | | | | 116 |
| Aker | 388 | 388 | 388 | | | | | 776 |
| Gloucester | 265 | 265 | 265 | | | | | 530 |
| VICTOR | 427 | 427 | 427 | | | | | 854 |
| Sum training | 1 138 | 1 138 | 1 138 | | | | | 2 276 |
| QUASAR 2 | 1 112 | 1 110 | 1 106 | 1 105 | 1 084 | 1 096 | | 5 501 |
| TransSCOT | 2 856 | 2 856 | 2 856 | 2 856 | | 2 547 | 2 534 | 2 856 | 16 505 |

**Supplementary Table 2 | Patient and scan count in LNM prediction**

| Cohort | Patients | Scans | | | |
|---|---|---|---|---|---|
| | | Aperio AT2 | NanoZoomer XR | NanoZoomer S210 | Sum |
| Ahus | 69 | 69 | 69 | | 138 |
| Aker | 243 | 243 | 243 | | 486 |
| Gloucester | 444 | 444 | 444 | | 888 |
| VICTOR | 282 | 282 | 282 | | 564 |
| Mainz | 135 | 500 | 500 | | 1 000 |
| DENEB | 61 | | | 61 | 61 |
| Sum training | 1 234 | 1 538 | 1 538 | 61 | 3 137 |
| Dutch T1 | 290 | 286 | 288 | | 574 |



\end{document}